\documentclass{article}

\usepackage{iclr2027_conference,times}

\usepackage[utf8]{inputenc}
\usepackage[T1]{fontenc}
\usepackage{url}
\usepackage{xurl}
\usepackage{booktabs}
\usepackage{amsfonts}
\usepackage{amsmath}
\usepackage{amssymb}
\usepackage{graphicx}
\usepackage{microtype}
\usepackage{xspace}
\usepackage{longtable}
\usepackage[table]{xcolor}
\usepackage{arydshln}
\usepackage{tikz}
\usetikzlibrary{arrows.meta,positioning}
\usepackage{float}
\usepackage{wrapfig}
\usepackage{algorithm}
\usepackage{algpseudocode}
\usepackage{hyperref}
\hypersetup{hidelinks}
\usepackage{placeins}
\usepackage{needspace}
\usepackage{capt-of}

\newcommand{\Ours}{\textsc{SearchSkill}\xspace}
\newcommand{\RepoLink}{\href{https://anonymous.4open.science/r/SearchSkill-BEEE}{here}\xspace}
\definecolor{skilltagcolor}{RGB}{74,170,255}
\definecolor{searchtagcolor}{RGB}{68,196,138}
\definecolor{answertagcolor}{RGB}{241,112,122}
\definecolor{infotagcolor}{RGB}{245,173,74}
\definecolor{selecttagcolor}{RGB}{128,111,214}
\definecolor{thinktagcolor}{RGB}{170,136,255}
\definecolor{skillgainbg}{RGB}{232,246,234}
\definecolor{deltabrown}{RGB}{236,184,113}
\definecolor{plotorange}{RGB}{247,190,132}
\definecolor{plotorangedark}{RGB}{190,84,36}
\definecolor{plotblue}{RGB}{176,209,216}
\definecolor{plotbluedark}{RGB}{42,118,143}
\definecolor{plotgrid}{RGB}{224,224,224}
\definecolor{rlgain}{RGB}{35,132,67}
\definecolor{rlloss}{RGB}{190,65,55}
\definecolor{ssBlue}{HTML}{E1EDF8}
\definecolor{ssPurple}{HTML}{ECE5F4}
\definecolor{ssPeach}{HTML}{F8E9DA}
\definecolor{ssGroup}{HTML}{F0F2F5}
\newcommand{\tagopen}[2]{\texttt{\textcolor{#1}{<#2>}}}
\newcommand{\tagclose}[2]{\texttt{\textcolor{#1}{</#2>}}}

\newcommand{\flipcell}[3]{#1 {\scriptsize(\textcolor{rlgain}{+#2}/\textcolor{rlloss}{-#3})}}
\title{\Ours: Teaching LLMs to Use Search Tools with Evolving Skill Banks}

\author{%
\parbox[t]{\dimexpr\textwidth-2\tabcolsep\relax}{%
\centering\normalfont
\textbf{Hu Jinchao}\textsuperscript{1}\quad
\textbf{Meizhi Zhong}\textsuperscript{2}\quad
\textbf{Kehai Chen}\textsuperscript{1}\quad
\textbf{Min Zhang}\textsuperscript{1}\\
\textsuperscript{1}School of Computer Science and Technology, Harbin Institute of Technology, Shenzhen\\
\textsuperscript{2}Independent Researcher\\
Shenzhen / Beijing, China\\
\texttt{jchu@stu.hit.edu.cn}\quad\texttt{meizhi.zhong.1999@gmail.com}\\
\texttt{chenkehai@hit.edu.cn}\quad\texttt{zhangmin2021@hit.edu.cn}%
}
}

\hypersetup{pdfauthor={Hu Jinchao, Meizhi Zhong, Kehai Chen, Min Zhang}}

\iclrfinalcopy

\begin{document}
\maketitle
\pagestyle{plain}
\vspace{-2\baselineskip}
\raggedbottom

\begin{abstract}
  Teaching language models to use search tools is not only a question of whether they search, but also of whether they issue good queries. This is especially important in open-domain question answering, where broad or copied queries often waste retrieval budget and derail later reasoning. We propose \Ours, a framework that makes query planning explicit through reusable search skills. At each step, the model first selects a skill, then generates a search or answer action conditioned on the selected skill card. The skill inventory itself is not fixed: SearchSkill maintains an evolving SkillBank, expands or refines it from recurrent failure patterns, and reconstructs affected trajectories before supervised training. The resulting two-stage SFT recipe aligns training with the inference-time protocol of skill selection followed by skill-grounded execution. Across open-source and closed-source models, SearchSkill improves exact match on knowledge-intensive QA benchmarks and yields better retrieval behavior, including fewer copied first queries, more atomic hop-focused queries, and more correct answers within a small search budget. These results suggest that explicit skill-conditioned query planning is a lightweight alternative to treating search as an undifferentiated action. All training code and configurations are available \RepoLink.
\end{abstract}

\section{Introduction}\label{sec:introduction}
Large language models (LLMs) exhibit strong reasoning ability and broad parametric knowledge, but that knowledge remains fundamentally static: it can be outdated, incomplete, or simply missing the long-tail facts required by knowledge-intensive tasks. This limitation is especially visible in open-domain and multi-hop question answering, where solving a problem often requires gathering fresh evidence, following hidden bridge entities, and verifying intermediate conclusions against external sources. Search tools therefore play an indispensable role in extending the knowledge boundary of LLMs and turning them into reliable systems for real-world information seeking.

Recent years have seen rapid progress on retrieval-augmented and tool-using LLMs. Retrieval-augmented generation (RAG) augments generation with external passages for knowledge-intensive prediction~\citep{lewis2021rag}. Tool-use methods such as Toolformer~\citep{schick2023toolformer} and ReAct~\citep{yao2023react} further enable models to interleave reasoning with external actions. More recent search-centered systems train this capability more directly: Search-R1~\citep{jin2025searchr1} optimizes long-horizon search behavior with reinforcement learning, while ZeroSearch~\citep{sun2025zerosearch} incentivizes search capability without relying on live search during training. These advances have substantially improved LLM--retriever interaction, but they also expose a practical bottleneck. 

\textbf{Challenge 1: most existing methods teach the model to search, but devote much less modeling capacity to how to formulate high-quality search queries.} In multi-hop settings, poor queries lead to longer interaction traces, redundant tool calls, noisy evidence, and lower final answer accuracy. The issue is often not whether a model uses a retriever, but whether it does so intelligently.

Recent work on agent skills offers a complementary angle. Anthropic's Agent Skills package instructions, scripts, and resources into reusable modules that agents can discover and load on demand~\citep{zhang2025agentskills}. Analyses of skill ecosystems show that skills rapidly became a practical mechanism for extending model functionality, while also raising ecosystem-level questions about organization and safe reuse~\citep{ling2026agentskillsdatadrivenanalysis,liu2026agentskillswildempirical}. AgentSkillOS studies selection and benchmarking over large skill ecosystems~\citep{li2026agentskillos}, while Reinforcement Learning for Self-Improving Agent with Skill Library, MemSkill, and SkillRL explore how agents can maintain and evolve skill libraries or skill banks over training~\citep{wang2026skilllibraryrl,zhang2026memskill,xia2026skillrl}. However, applying this intuition to search tool use is still non-trivial. 

\textbf{Challenge 2: current skill frameworks do not directly tell us how skills should be represented, invoked, and updated inside search trajectories.} A useful search skill must shape not only a high-level strategy, but also the next query, the arguments passed to the retriever, the verification behavior after retrieval, and the stopping decision before answering.

To address these challenges, we propose \Ours, a framework that teaches LLMs to use search tools through an evolving bank of reusable search skills. Each interaction follows \emph{select--read--act}: the model selects skill ids, reads their procedural cards, and emits an explicit skill declaration followed by a search query or final answer. Retrieved evidence updates the context for the next decision. This token-level interface makes skill selection and execution explicit, trainable parts of the model's search trajectory. The cards describe reusable operations such as resolving a bridge entity, comparing attributes, checking conflicting evidence, and extracting a grounded answer.

The same interface connects skill discovery to policy training. A skill generator proposes and revises cards using retrieval experience, retaining procedures supported by successful rollouts. The updated bank guides trajectory reconstruction, and two-stage SFT teaches the policy to select and execute its procedures. The target backbone or a stronger model can serve as the skill generator. With Qwen-generated skills, the 7B policy reaches 38.60 macro EM, comparable to Search-R1's 38.68, while the 3B policy reaches 34.87, as reported in Table~\ref{tab:main-results}. Using GPT-5.4 as the skill generator yields the best 7B and 3B scores of 42.34 and 36.94.

\begin{samepage}
Our main contributions are as follows:
\begin{itemize}
  \item We propose a skill-conditioned search-tool interface that improves query formulation by making the model choose a reusable search strategy before each tool call, addressing the common failure of issuing broad or copied multi-hop queries.
  \item We introduce an evolvable Search SkillBank that is updated through failure analysis, add/refine operations, and trajectory reconstruction, providing a reusable tool-use prior that can extend beyond the QA setting.
  \item We show that Qwen-generated skills and two-stage SFT yield performance comparable to RL-trained search agents, suggesting that self-directed skill discovery can elicit the model's latent capabilities and enable more efficient and accurate search.
  \item We demonstrate effectiveness on both open-source and closed-source models: fine-tuned Qwen2.5 3B/7B models improve QA performance, and the frozen SkillBank also benefits proprietary models as an external planning guide.
\end{itemize}
\end{samepage}

\section{Preliminaries}
\label{sec:preliminaries}

\subsection{Agentic search}
Agentic search is a sequential decision process in which a language model interleaves reasoning with search and uses retrieved information to guide subsequent actions~\citep{yao2023react,jin2025searchr1}.

Formally, given a question \(q\), a policy \(\pi_\theta\) chooses an action \(a_t\) conditioned on the interaction history \(h_t\), which records \(q\), prior actions, and retrieved passages. A search action issues a query \(u_t\) and receives passages \(e_t=\mathcal{R}(u_t)\) from a retriever \(\mathcal{R}\); an answer action returns \(y\) and ends the interaction. After a search, \(h_{t+1}\) includes the query and returned passages, allowing subsequent actions to depend on the newly retrieved information.

The objective is to answer correctly within a retrieval budget \(T\), where each search consumes one unit. For multi-hop questions, retrieved passages may establish intermediate facts without resolving the entire question. The policy must therefore track resolved facts and remaining information needs, using earlier findings to determine subsequent retrieval targets. It must also decide when the available information is sufficient to answer, since redundant or uninformative queries consume the budget.

\newpage
\subsection{Search skills and the SkillBank}
A skill is a reusable procedure specifying when and how to perform an operation~\citep{zheng2025skillweaver,wang2026skilllibraryrl,zhang2026memskill,xia2026skillrl}. We represent a \emph{search skill} by an identifier \(i\) and a natural-language card \(c_i\). The card states the conditions for applying the skill, its procedure for querying and using retrieved information, the intermediate facts to establish, and when to continue searching or produce an answer.

A \emph{SkillBank} \(B=\{(i,c_i)\}_{i\in\mathcal I_B}\) is a collection of such skills, where \(\mathcal I_B\) denotes the available identifiers. Its compact index \(\operatorname{Index}(B)\) summarizes the skills for selection, while the corresponding cards provide detailed instructions for execution. This representation allows a search procedure to be reused across questions.

\subsection{Skill-conditioned interaction}
SearchSkill organizes each decision through a \emph{select--read--act} interface, illustrated by the inference example in Figure~\ref{fig:searchskill-overview}. Given \(h_t\), the policy selects one or more skill identifiers \(s_t\subseteq\mathcal I_B\) from \(\operatorname{Index}(B)\). The system loads their cards \(C_B(s_t)\), and the policy uses these instructions to generate a search or answer action \(a_t\). The joint probability factorizes as
\begin{equation}
 p_\theta(s_t,a_t\mid h_t,B)
 =p_\theta(s_t\mid h_t,\operatorname{Index}(B))\,
 p_\theta(a_t\mid h_t,s_t,C_B(s_t)).
 \label{eq:skill-factorization}
\end{equation}
The first factor models skill selection; the second models action generation conditioned on the selected cards. Search actions and retrieved passages update the history for the next selection; an answer ends the interaction. Section~\ref{sec:searchskill} describes how the bank and policy are learned; Appendix~\ref{app:related-work} provides additional related work.

\section{Methodology}\label{sec:searchskill}
\begin{figure}[t]
  \centering
  \includegraphics[width=\linewidth]{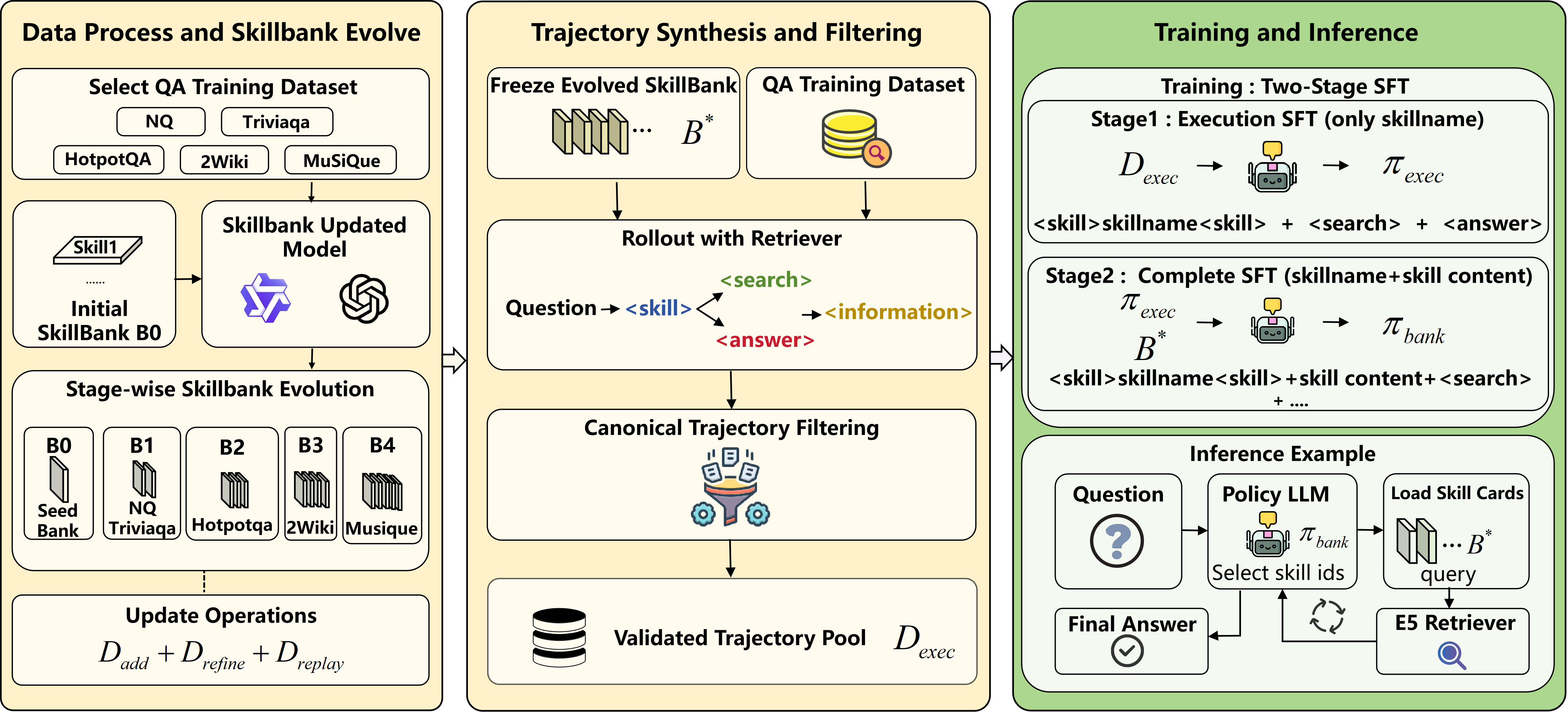}
  \caption{\textbf{Overview of \Ours.} A skill generator builds and refines the SkillBank through retrieval feedback. The evolved bank guides trajectory construction, and two-stage SFT teaches skill execution and skill selection. Given a question, the trained policy selects skills, reads their cards, and gathers evidence through retrieval before producing a final answer.}
  \label{fig:searchskill-overview}
\end{figure}

As shown in Figure~\ref{fig:searchskill-overview}, SearchSkill first evolves a SkillBank from retrieval experience, then constructs skill-guided trajectories and trains the policy in two supervised stages. Bank evolution updates the cards while model weights remain fixed; SFT updates the policy while the final bank remains fixed.

\subsection{Stage-wise SkillBank evolution}
We begin from a compact seed bank \(B_0\) containing six general retrieval and grounding skills, and expand it through four rounds: \(B_1\) from NQ and TriviaQA, \(B_2\) from HotpotQA, \(B_3\) from 2WikiMultiHopQA, and \(B_4\) from MuSiQue. This ordering establishes direct lookup before extending the bank to bridge resolution, comparison, and longer reasoning chains.

\textbf{Construct candidate skills.} In each round, we group training questions using available dataset metadata and model-based grouping to preserve coverage. Representative question--answer pairs and the current bank are provided to a skill generator \(\mathcal G\). The generator identifies recurring search procedures that existing cards do not cover, then proposes new cards or refines existing ones. Skill identifiers and procedures are generated from the examples; the grouping labels serve only to organize coverage. Each card specifies an observable trigger, retrieval steps, facts to establish, failure modes to avoid, and conditions for continuing or answering. This yields a candidate bank.

\textbf{Test and refine.} The generator uses the candidate cards to search and answer training questions. Reference answers are withheld from the rollout context and used to score its final answer. Exact match identifies successful trajectories; the skills used in them are retained and protected. Protection preserves their identifiers and core procedures in later updates. Aggregate outcomes and representative success and failure traces are returned to \(\mathcal G\), which can revise, merge, or remove unprotected cards. The revised bank enters the next round, carrying forward previously validated skills. After four rounds, we fix \(B^\star=B_4\) for trajectory construction and training. Grouping procedures, card formats, and update details appear in Appendices~\ref{app:self-generation} and~\ref{app:skillbank-evolution}.

\subsection{Skill-guided trajectory construction}
With \(B^\star\) fixed, we sample questions across the five training datasets and generate trajectories through interaction with the retriever. At each step, the rollout model predicts a primary skill, optional supporting skills, a search or answer action, and a checkpoint describing the resolved intermediate state. The retriever executes search queries, and returned passages are appended to the history until the model produces a final answer.

We keep only canonical trajectories that satisfy answer correctness, at least one retrieval step, legal skill identifiers, and consistency between the declared skill and executed action. These form the trajectory set \(\mathcal D_{\mathrm{exec}}\), with explicit \texttt{<skill>} declarations before \texttt{<search>} or \texttt{<answer>} actions. Appendix~\ref{app:trajectory-construction} details data construction.

\subsection{Two-stage supervised fine-tuning}
\textbf{Stage I: skill execution.} The first stage starts from the target backbone and teaches the basic SearchSkill action format: each assistant turn emits \texttt{<skill>} followed by \texttt{<search>} or \texttt{<answer>}. For a trajectory \(\tau\), let \(h_t\) be the context before action turn \(a_t\), and \(\mathcal M(\tau)\) index the supervised steps. We optimize
\begin{equation}
\mathcal L_{\mathrm{exec}}(\theta)
=-\sum_{\tau\in\mathcal D_{\mathrm{exec}}}\sum_{t\in\mathcal M(\tau)}
\lambda(a_t)\sum_{k=1}^{|a_t|}
\log p_\theta(a_{t,k}\mid h_t,a_{t,<k}),
\label{eq:sft-exec}
\end{equation}
where \(\lambda\) weights the targets. Full-trajectory records supervise all assistant turns; closure-focused records retain the preceding interaction as context but supervise only the final answer. This stage teaches skill-tagged query planning, evidence use, and stopping behavior.

\textbf{Stage II: skill selection.} The second stage continues from the first-stage checkpoint and repacks the same trajectory supervision into the select--read--act protocol. Each original action is rewritten as a selection turn \(s_t\), a user message containing the selected cards \(C_{B^\star}(s_t)\), and the original action turn \(a_t\). The policy selects from the bank index without a question-specific recommended-skill hint. Let \(\mathcal D_{\mathrm{bank}}=\mathcal R_{B^\star}(\mathcal D_{\mathrm{exec}})\) denote the rewritten dataset, \(h_t^{\mathrm{sel}}\) the context before selection, and \(h_t^{\mathrm{act}}=h_t^{\mathrm{sel}}\oplus s_t\oplus C_{B^\star}(s_t)\) the context after card insertion, where \(\oplus\) concatenates messages. We~optimize
\begin{equation}
\begin{aligned}
\mathcal L_{\mathrm{bank}}(\theta)
=-\sum_{\tau'\in\mathcal D_{\mathrm{bank}}}\sum_{t\in\mathcal M(\tau')}
\Bigg[&\lambda(s_t)\sum_{k=1}^{|s_t|}
\log p_\theta(s_{t,k}\mid h_t^{\mathrm{sel}},s_{t,<k})\\
&+\lambda(a_t)\sum_{k=1}^{|a_t|}
\log p_\theta(a_{t,k}\mid h_t^{\mathrm{act}},a_{t,<k})\Bigg].
\end{aligned}
\label{eq:sft-bank}
\end{equation}
Here \(\lambda\) is zero for unsupervised targets; supplied cards and retrieved passages are context only. Thus the first objective teaches execution, while the second teaches selection and card-conditioned execution under the inference-time protocol. At inference, both the policy and bank are fixed. Algorithm~\ref{alg:searchskill-rollout} gives the full interaction loop, and Appendix~\ref{app:two-stage-sft-training} lists training settings.

\section{Experiments}\label{sec:experiments}
\subsection{Experimental setup}\label{sec:experimental-setup}
\textbf{Datasets.}
We evaluate on seven QA benchmarks: Natural Questions (NQ), TriviaQA, and PopQA for single-hop retrieval~\citep{kwiatkowski2019naturalquestions,joshi2017triviaqa,mallen2023popqa}; HotpotQA, 2WikiMultiHopQA, MuSiQue, and Bamboogle for multi-hop reasoning~\citep{yang2018hotpotqa,ho2020wikimultihopqa,trivedi2022musique,press2023bamboogle}. Together, they test factual lookup and the ability to decompose questions requiring multiple retrieval steps.

\textbf{Baselines.}
We compare with direct inference, chain-of-thought prompting~\citep{wei2022cot}, RAG~\citep{lewis2021rag}, and Search-o1~\citep{li2025searcho1}, as well as RL-based search agents: ZeroSearch~\citep{sun2025zerosearch}, Search-R1~\citep{jin2025searchr1}, StepSearch~\citep{wang2025stepsearch}, Smart-Searcher (R1-Searcher++)~\citep{song2025r1searcherplus}, and s3~\citep{jiang2025s3}. These baselines compare language-only reasoning, retrieval access, and methods trained to search.

\textbf{Models and training.} We use Qwen2.5-7B and Qwen2.5-3B~\citep{qwen2025qwen25}, with base and instruct variants, to examine both model scale and initialization. The labels \emph{7B skills}, \emph{3B skills}, and \emph{GPT-5.4 skills} identify the model that evolves the SkillBank and generates training trajectories through the workflow in Section~\ref{sec:searchskill}. Each policy is trained with two-stage SFT and evaluated with its frozen bank.

Training questions come from the training splits of NQ, TriviaQA, HotpotQA, 2WikiMultiHopQA, and MuSiQue. Sampling preserves coverage across question types and retrieval demands. Successful trajectories provide SFT targets, while failed trajectories guide skill revision and are excluded from the SFT targets. Appendices~\ref{app:self-generation} and~\ref{app:experimental-setups} give data construction and training details.

\textbf{Evaluation.} Search-based methods share an E5 retriever~\citep{wang2022e5} returning the top-3 passages per query, a five-search budget, and greedy decoding. Retrieved passages are appended to the interaction history for subsequent decisions. We report normalized exact match (EM) as a percentage; \emph{Avg.} is the unweighted mean over seven datasets. Our main \Ours results use SFT alone; post-SFT GRPO is analyzed separately in Section~\ref{sec:further-analysis-rl}.

\begin{table}[t]
  \centering
  \caption{Exact match percentages on seven QA benchmarks. Avg. is their unweighted mean. Within each model scale, best results are bold and second-best results are underlined. Parentheses identify the skill generator.}
  \label{tab:main-results}
  \setlength{\tabcolsep}{3pt}
  \renewcommand{\arraystretch}{1.12}
  \resizebox{\textwidth}{!}{%
  \begin{tabular}{lcccccccc}
    \toprule
    & \multicolumn{3}{c}{\cellcolor{ssBlue}\textbf{Single-Hop QA}} & \multicolumn{4}{c}{\cellcolor{ssPurple}\textbf{Multi-Hop QA}} & \cellcolor{ssPeach} \\
    \textbf{Method} &
    \cellcolor{ssBlue}\textbf{NQ} &
    \cellcolor{ssBlue}\textbf{TriviaQA} &
    \cellcolor{ssBlue}\textbf{PopQA} &
    \cellcolor{ssPurple}\textbf{HotpotQA} &
    \cellcolor{ssPurple}\textbf{2Wiki} &
    \cellcolor{ssPurple}\textbf{MuSiQue} &
    \cellcolor{ssPurple}\textbf{Bamboogle} &
    \cellcolor{ssPeach}\textbf{Avg. (\(\uparrow\))} \\
    \midrule
    \addlinespace[0.25em]
    \multicolumn{9}{l}{\cellcolor{ssGroup}\textbf{Qwen2.5-7B-Base/Instruct}} \\
    \addlinespace[0.15em]
    Direct Inference & 16.79 & 38.98 & 15.36 & 17.00 & 26.00 & 1.00 & 8.00 & 17.59 \\
    CoT & 12.19 & 34.03 & 16.11 & 17.00 & 18.50 & 5.00 & 22.40 & 17.89 \\
    RAG & 38.00 & 58.01 & 39.10 & 33.50 & 23.00 & 7.00 & 8.80 & 29.63 \\
    Search-o1 & 30.94 & 57.20 & 36.02 & 31.50 & 27.50 & 11.00 & 36.80 & 32.99 \\
    ZeroSearch-Base & 36.70 & 58.90 & 47.30 & 29.00 & 31.00 & 7.50 & 22.40 & 33.26 \\
    SearchR1-Base (GRPO) & 39.50 & 56.00 & 38.80 & 35.50 & 40.50 & 9.00 & 37.60 & 36.70 \\
    SearchR1-Instruct (GRPO) & 39.66 & \underline{60.51} & 41.09 & 40.00 & 37.50 & \textbf{16.00} & 36.00 & 38.68 \\
    StepSearch & 33.21 & 52.49 & 45.90 & 37.00 & 40.00 & 13.50 & \textbf{44.80} & 38.13 \\
    Smart-Searcher (R1-Searcher++) & 38.42 & 59.65 & 47.20 & 37.50 & 37.50 & 13.50 & \underline{44.00} & 39.68 \\
    s3 & 6.65 & 30.08 & 17.10 & 21.50 & 14.50 & 4.00 & 11.20 & 15.00 \\
    \midrule
    \Ours-Instruct (7B skills) & 37.60 & 60.00 & 47.40 & 39.50 & 40.50 & 14.00 & 31.20 & 38.60 \\
    \Ours-Base (GPT-5.4 skills) & \underline{40.89} & 60.01 & \textbf{52.80} & \underline{42.00} & \textbf{47.00} & 14.50 & 39.20 & \textbf{42.34} \\
    \Ours-Instruct (GPT-5.4 skills) & \textbf{41.60} & \textbf{61.04} & \underline{51.40} & \textbf{43.00} & \underline{44.50} & \underline{15.50} & 38.40 & \underline{42.21} \\
    \midrule
    \addlinespace[0.25em]
    \multicolumn{9}{l}{\cellcolor{ssGroup}\textbf{Qwen2.5-3B-Base/Instruct}} \\
    \addlinespace[0.15em]
    Direct Inference & 11.88 & 33.31 & 11.64 & 14.50 & 27.00 & 2.00 & 4.00 & 14.90 \\
    CoT & 11.33 & 33.07 & 11.90 & 15.00 & 15.50 & 2.50 & 20.80 & 15.73 \\
    RAG & 33.16 & 53.24 & 38.30 & 27.50 & 26.00 & 3.00 & 6.40 & 26.80 \\
    Search-o1 & 28.86 & 47.49 & 31.29 & 24.50 & 24.00 & 6.00 & 20.80 & 26.13 \\
    ZeroSearch-Base & 34.00 & \underline{57.50} & 41.40 & 26.00 & 30.00 & 6.50 & 16.00 & 30.20 \\
    SearchR1-Base (GRPO) & 36.10 & \textbf{58.00} & 41.30 & 33.00 & 31.50 & 7.50 & 14.40 & 31.69 \\
    SearchR1-Instruct (GRPO) & 33.60 & 56.00 & 43.90 & 29.00 & 31.00 & 10.50 & \textbf{28.00} & 33.14 \\
    \midrule
    \Ours-Instruct (3B skills) & 35.80 & 53.90 & 43.90 & \textbf{39.50} & 35.50 & 11.50 & 24.00 & 34.87 \\
    \Ours-Base (GPT-5.4 skills) & \underline{36.20} & 53.90 & \textbf{47.00} & 34.00 & \underline{37.50} & \underline{13.50} & \textbf{28.00} & \underline{35.73} \\
    \Ours-Instruct (GPT-5.4 skills) & \textbf{37.50} & 54.30 & \underline{46.60} & \underline{35.50} & \textbf{41.50} & \textbf{16.00} & \underline{27.20} & \textbf{36.94} \\
    \bottomrule
  \end{tabular}}
\end{table}


\begin{samepage}
\subsection{Main results}
\label{sec:main-results}
\textbf{Self-generated skills elicit competitive search capabilities.} Table~\ref{tab:main-results} shows that \Ours with Qwen-generated skills reaches 38.60 macro EM on 7B, comparable to Search-R1 at 38.68 and exceeding StepSearch at 38.13. The 3B model also reaches 34.87, surpassing Search-R1-Instruct by 1.73 points. These results highlight the potential of self-directed skill discovery: by discovering and refining its own search procedures and learning to apply them through two-stage SFT, a 7B model can develop search capabilities competitive with those obtained through large-scale reinforcement learning.
\looseness=-1\par
\end{samepage}

\textbf{Stronger skill generators further improve performance.} With GPT-5.4 as the skill generator, \Ours obtains the best macro averages in both blocks, reaching 42.34 on 7B and 36.94 on 3B. Compared with the strongest RL baseline in each block, the average gains are 2.66 and 3.80 points. \textbf{Search skills yield substantial gains on multi-hop QA.} For example, on 2Wiki, where the model must decompose relations and carry intermediate entities across searches, 7B-Base and 3B-Instruct improve over their corresponding Search-R1 variants by 6.50 and 10.50 points, respectively. \textbf{Single-hop QA is not sacrificed for this gain}: with only SkillBank-guided SFT, \Ours remains on par with or better than RL-trained search agents on factual lookup benchmarks, while avoiding their large online rollout cost. \textbf{The effect is stable across scale and initialization}: both 7B and 3B blocks reach the best macro average, and the gains appear for base and instruct variants, indicating that the evolved SkillBank provides a reusable search-control prior rather than a model-specific prompt advantage. These results suggest that stronger skill generators offer a practical route to strong search performance without expensive, time-consuming training.

\Needspace{15\baselineskip}
\subsection{Generalization to web-search tasks}
\label{sec:web-transfer}

\begingroup
\setlength{\intextsep}{4pt}
\begin{wraptable}{r}{0.46\textwidth}
  \centering
  \small
  \setlength{\abovecaptionskip}{0pt}
  \setlength{\belowcaptionskip}{6pt}
  \caption{GAIA accuracy in percent. Avg. averages both levels.}
  \label{tab:gaia-transfer}
  \setlength{\tabcolsep}{0pt}
  \renewcommand{\arraystretch}{1.15}
  \begin{tabular*}{\linewidth}{@{\extracolsep{\fill}}lrrr@{}}
    \toprule
    Method & Level 1 & Level 2 & Avg. \\
    \midrule
    Base & 9.52 & 0.00 & 4.76 \\
    Search-R1 & 16.67 & 4.55 & 10.61 \\
    \Ours & 11.90 & \textbf{6.06} & 8.98 \\
    \bottomrule
  \end{tabular*}
\end{wraptable}
\textbf{Generalization of QA skills.} In our evaluation on GAIA~\citep{mialon2023gaia} Level~1 and Level~2, we deploy the QA-trained 7B policy with its existing SkillBank and Serper web search, without additional training. As shown in Table~\ref{tab:gaia-transfer}, \Ours approaches Search-R1's average accuracy across the two levels and performs better on the more challenging Level~2, reaching 6.06\% versus 4.55\%. These results show that QA-derived skills generalize to evidence gathering in unseen web-search environments.
\par
\endgroup

\begingroup
\setlength{\intextsep}{4pt}
\begin{wraptable}{r}{0.50\textwidth}
  \centering
  \small
  \setlength{\abovecaptionskip}{0pt}
  \setlength{\belowcaptionskip}{6pt}
  \caption{WebWalkerQA multi-source accuracy in percent.}
  \label{tab:webwalker-transfer}
  \setlength{\tabcolsep}{0pt}
  \renewcommand{\arraystretch}{1.15}
  \begin{tabular*}{\linewidth}{@{\extracolsep{\fill}}lrrrr@{}}
    \toprule
    Method & Easy & Medium & Hard & Overall \\
    \midrule
    Base & 8.75 & 6.43 & 4.17 & 6.18 \\
    Reflexion & 13.75 & 6.43 & 2.50 & 6.76 \\
    \Ours & 12.50 & \textbf{8.57} & \textbf{6.67} & \textbf{8.82} \\
    \bottomrule
  \end{tabular*}
\end{wraptable}
\textbf{Extending the skill-learning workflow.} Starting from a seed SkillBank \(B_0\), Qwen2.5-7B-Instruct evolves its own SkillBank for WebWalkerQA~\citep{wu2025webwalker}. We then apply the same two-stage SFT procedure using 200 validated trajectories and evaluate 340 multi-source questions with a 32K context and 15 additional webpage visits. As shown in Table~\ref{tab:webwalker-transfer}, \Ours outperforms Base and Reflexion~\citep{shinn2023reflexion} on Medium and Hard questions and achieves the highest overall accuracy at 8.82\%. These results demonstrate that the same workflow can be reapplied to a new web-search task, enabling a 7B model to develop effective skills for navigating and integrating evidence across pages.
\par
\endgroup

\Needspace{10\baselineskip}
\subsection{SkillBank ablations}
\textbf{SkillBank information and evolution.}
\label{sec:further-analysis-skillbank-content}

\begin{figure}[t]
  \centering
  \begin{minipage}[t]{0.47\linewidth}
    \vspace{0pt}
    \centering
    \includegraphics[width=\linewidth]{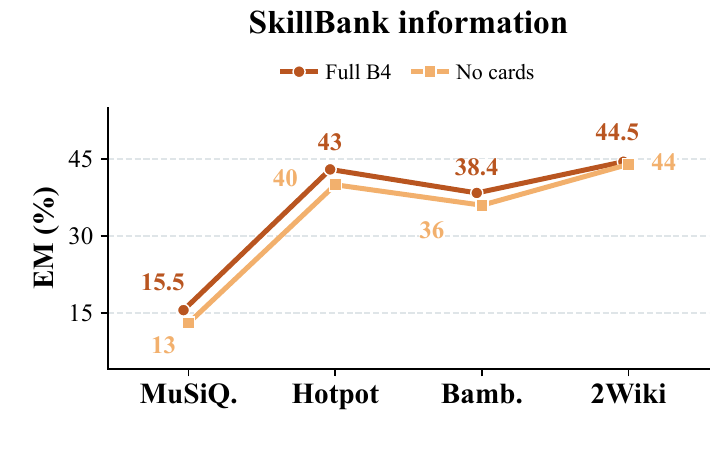}
  \end{minipage}
  \hfill
  \begin{minipage}[t]{0.47\linewidth}
    \vspace{0pt}
    \centering
    \includegraphics[width=\linewidth]{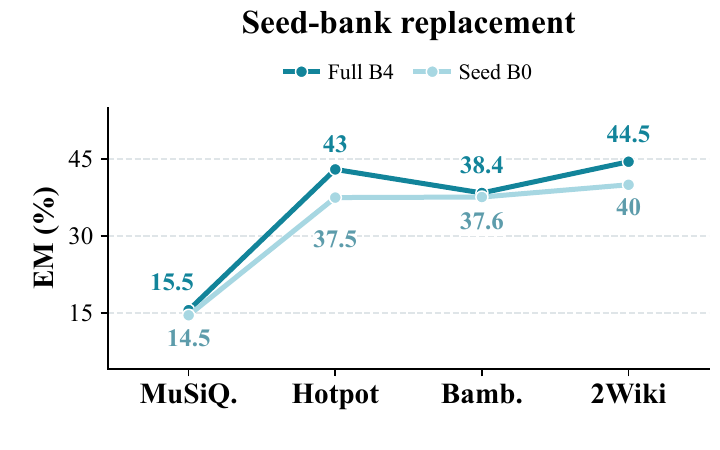}
  \end{minipage}
  \caption{SkillBank information and evolution under the same SFT policy. Left: full \(B_4\) versus skill IDs without cards. Right: evolved \(B_4\) versus seed \(B_0\).}
  \label{fig:skillbank-content-init}
\end{figure}

Figure~\ref{fig:skillbank-content-init} separates two sources of SkillBank benefit. The left plot keeps the selected \(B_4\) ids but removes their card content (\emph{No cards}), testing whether names alone are sufficient. The right plot keeps the same SFT policy but replaces the evolved \(B_4\) with the seed bank \(B_0\), testing the effect of bank evolution. Both controls generally reduce performance, showing that SearchSkill benefits from both readable skill-card content and iterative SkillBank refinement. The drops are not catastrophic because the trained backbone already acquires a basic query-decomposition prior from trajectory SFT, but explicit skill content and evolved routing still make the decomposition more reliable.

\begin{figure}[!t]
  \centering
  \begin{minipage}[t]{0.34\linewidth}
    \centering
      \includegraphics[width=\linewidth]{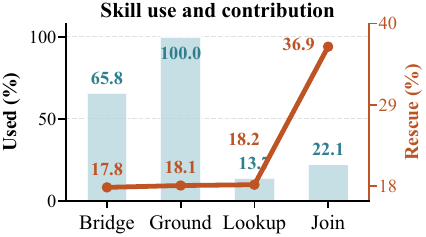}
  \end{minipage}
  \hfill
  \begin{minipage}[t]{0.205\linewidth}
    \centering
      \includegraphics[width=\linewidth]{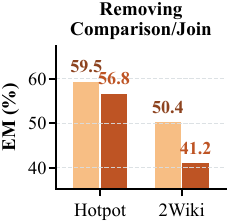}
  \end{minipage}
  \hfill
  \begin{minipage}[t]{0.205\linewidth}
    \centering
      \includegraphics[width=\linewidth]{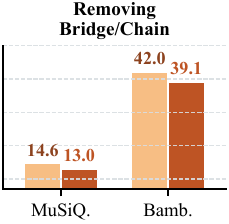}
  \end{minipage}
  \hfill
  \begin{minipage}[t]{0.205\linewidth}
    \centering
      \includegraphics[width=\linewidth]{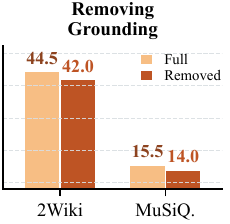}
  \end{minipage}
  \caption{Skill use and contribution. Left: category activation and Rescue rates. Right: EM with the full bank (light) and after removing a skill category (dark).}
  \label{fig:skillbank-evolution}
  \vspace{6pt}
  \begingroup
  \setlength{\abovecaptionskip}{5pt}
  \setlength{\belowcaptionskip}{4pt}
  \captionof{table}{GPT-5.4 distillation control on Qwen2.5-7B-Instruct (EM, \%).}
  \label{tab:teacher-distillation}
  \small
  \setlength{\tabcolsep}{3pt}
  \renewcommand{\arraystretch}{1.05}
  \begin{tabular*}{\linewidth}{@{\extracolsep{\fill}}lrrrrrrrr@{}}
    \toprule
    Method & NQ & TriviaQA & PopQA & HotpotQA & 2Wiki & MuSiQue & Bamboo. & Avg. \\
    \midrule
    \Ours & 41.60 & 61.04 & 51.40 & 43.00 & 44.50 & 15.50 & 38.40 & \textbf{42.21} \\
    GPT-5.4 trajectories & 38.64 & 58.20 & 46.60 & 38.00 & 32.50 & 18.00 & 33.60 & \textbf{37.93} \\
    \bottomrule
  \end{tabular*}
  \par
  \endgroup
\end{figure}

\textbf{Skill contribution after activation.}
\label{sec:further-analysis-skillbank-evolution}
\par\nopagebreak[4]
We evaluate whether activated skills are functionally necessary rather than merely selected. Figure~\ref{fig:skillbank-evolution} groups \(B_4\) cards into direct lookup, bridge/chain, comparison/join, and grounding/verification. \emph{Activation} measures category use over multi-hop questions; \emph{Rescue} is the percentage of activated questions answered correctly and judged to require that category. Removing one category at a time generally reduces EM, showing its contribution to answer construction. Appendix~\ref{app:diagnostic-metrics} gives detailed definitions and scoring rules.
\par

\subsection{Beyond teacher distillation}
\label{sec:teacher-distillation}
To test whether our gains come solely from teacher distillation, we ask GPT-5.4 to generate new retrieval trajectories without a SkillBank or skill actions, then fine-tune Qwen2.5-7B-Instruct with matched training questions, data size, and two-stage SFT budget. Table~\ref{tab:teacher-distillation} shows that this control reaches 37.93 macro EM, compared with 42.21 for \Ours, indicating that skill conditioning contributes beyond teacher supervision alone. Together with Section~\ref{sec:main-results}, these results suggest that using stronger models to develop reusable skills offers a low-cost, broadly transferable approach to improving search performance and enabling more efficient search in practical applications.
\par

\subsection{Query planning and evidence efficiency}
\label{sec:further-analysis-query-efficiency}
We further test whether \Ours addresses the query-quality failure that motivates our method: search policies often send the whole multi-hop question to the retriever instead of isolating the next missing entity or relation.
\par

\begingroup
\looseness=-1
Figure~\ref{fig:query-planning-efficiency} reports four diagnostics macro-averaged over HotpotQA, 2WikiMultiHopQA, MuSiQue, and Bamboogle. \emph{First query copy} (\(\downarrow\)) measures whether the first search is a near copy of the original question; \emph{atomic-hop query} (\(\uparrow\)) measures the fraction of searches that target one clear retrieval subgoal; \emph{average search} (\(\downarrow\)) counts the number of retrieval calls per example; and \emph{Correct@\(\leq3\) searches} (\(\uparrow\)) measures the fraction of examples answered correctly within three retrieval calls. Arrows indicate the preferred direction for each metric. See Appendix~\ref{app:diagnostic-metrics} for details.
\par\endgroup

\begin{figure}[t]
  \centering
  \setlength{\fboxsep}{0pt}
  \begin{minipage}[t]{0.24\linewidth}
    \centering
    \hspace*{1.8em}{\scriptsize\bfseries First Query Copy}\\[-0.1em]
    \begin{tikzpicture}[x=0.52cm,y=0.075cm]
      \foreach \y in {70,80,90,100} {
        \draw[plotgrid,dotted] (0,\y) -- (4.15,\y);
        \node[left,font=\tiny] at (0,\y) {\y};
      }
      \draw[black,thick] (0,70) rectangle (4.15,100);
      \draw[fill=plotorange!78,draw=none] (0.55,70) rectangle (1.15,89.93);
      \draw[fill=plotorange,draw=none] (1.78,70) rectangle (2.38,93.52);
      \draw[fill=plotorangedark,draw=none] (3.01,70) rectangle (3.61,76.00);
      \node[font=\tiny,plotorangedark] at (0.85,91.8) {89.9};
      \node[font=\tiny,plotorangedark] at (2.08,95.4) {93.5};
      \node[font=\tiny,plotorangedark] at (3.31,77.9) {76.0};
      \node[rotate=90,font=\scriptsize\bfseries] at (-1.18,85) {Copy (\%)};
      \node[rotate=25,anchor=east,font=\tiny] at (1.17,66.8) {w/o Skills};
      \node[rotate=25,anchor=east,font=\tiny] at (2.40,66.8) {SearchR1};
      \node[rotate=25,anchor=east,font=\tiny] at (3.63,66.8) {\Ours};
    \end{tikzpicture}
  \end{minipage}
  \hfill
  \begin{minipage}[t]{0.24\linewidth}
    \centering
    \hspace*{1.8em}{\scriptsize\bfseries Atomic Hop Query}\\[-0.1em]
    \begin{tikzpicture}[x=0.52cm,y=0.075cm]
      \foreach \y in {30,40,50,60} {
        \draw[plotgrid,dotted] (0,\y) -- (4.15,\y);
        \node[left,font=\tiny] at (0,\y) {\y};
      }
      \draw[black,thick] (0,30) rectangle (4.15,60);
      \draw[fill=plotblue!78,draw=none] (0.55,30) rectangle (1.15,32.88);
      \draw[fill=plotblue,draw=none] (1.78,30) rectangle (2.38,47.65);
      \draw[fill=plotbluedark,draw=none] (3.01,30) rectangle (3.61,50.02);
      \node[font=\tiny,plotbluedark] at (0.85,35.0) {32.9};
      \node[font=\tiny,plotbluedark] at (2.08,49.8) {47.7};
      \node[font=\tiny,plotbluedark] at (3.31,52.2) {50.0};
      \node[rotate=90,font=\scriptsize\bfseries] at (-1.18,45) {Atomic (\%)};
      \node[rotate=25,anchor=east,font=\tiny] at (1.17,26.6) {w/o Skills};
      \node[rotate=25,anchor=east,font=\tiny] at (2.40,26.6) {SearchR1};
      \node[rotate=25,anchor=east,font=\tiny] at (3.63,26.6) {\Ours};
    \end{tikzpicture}
  \end{minipage}
  \hfill
  \begin{minipage}[t]{0.24\linewidth}
    \centering
    \hspace*{1.8em}{\scriptsize\bfseries Average Search}\\[-0.1em]
    \begin{tikzpicture}[x=0.52cm,y=0.5625cm]
      \foreach \y in {1,2,3,4,5} {
        \draw[plotgrid,dotted] (0,\y) -- (4.15,\y);
        \node[left,font=\tiny] at (0,\y) {\y};
      }
      \draw[black,thick] (0,1) rectangle (4.15,5);
      \draw[fill=plotorange!78,draw=none] (0.55,1) rectangle (1.15,1.82);
      \draw[fill=plotorange,draw=none] (1.78,1) rectangle (2.38,4.12);
      \draw[fill=plotorangedark,draw=none] (3.01,1) rectangle (3.61,2.98);
      \node[font=\tiny,plotorangedark] at (0.85,2.08) {1.82};
      \node[font=\tiny,plotorangedark] at (2.08,4.38) {4.12};
      \node[font=\tiny,plotorangedark] at (3.31,3.24) {2.98};
      \node[rotate=90,font=\scriptsize\bfseries] at (-1.18,3) {Searches};
      \node[rotate=25,anchor=east,font=\tiny] at (1.17,0.56) {w/o Skills};
      \node[rotate=25,anchor=east,font=\tiny] at (2.40,0.56) {SearchR1};
      \node[rotate=25,anchor=east,font=\tiny] at (3.63,0.56) {\Ours};
    \end{tikzpicture}
  \end{minipage}
  \hfill
  \begin{minipage}[t]{0.24\linewidth}
    \centering
    \hspace*{1.8em}{\scriptsize\bfseries Correct@\( \le 3 \) Searches}\\[-0.1em]
    \begin{tikzpicture}[x=0.52cm,y=0.1125cm]
      \foreach \y in {10,15,20,25,30} {
        \draw[plotgrid,dotted] (0,\y) -- (4.15,\y);
        \node[left,font=\tiny] at (0,\y) {\y};
      }
      \draw[black,thick] (0,10) rectangle (4.15,30);
      \draw[fill=plotblue!78,draw=none] (0.55,10) rectangle (1.15,14.07);
      \draw[fill=plotblue,draw=none] (1.78,10) rectangle (2.38,21.79);
      \draw[fill=plotbluedark,draw=none] (3.01,10) rectangle (3.61,24.41);
      \node[font=\tiny,plotbluedark] at (0.85,15.6) {14.1};
      \node[font=\tiny,plotbluedark] at (2.08,23.3) {21.8};
      \node[font=\tiny,plotbluedark] at (3.31,25.9) {24.4};
      \node[rotate=90,font=\scriptsize\bfseries] at (-1.18,20) {Correct (\%)};
      \node[rotate=25,anchor=east,font=\tiny] at (1.17,7.8) {w/o Skills};
      \node[rotate=25,anchor=east,font=\tiny] at (2.40,7.8) {SearchR1};
      \node[rotate=25,anchor=east,font=\tiny] at (3.63,7.8) {\Ours};
    \end{tikzpicture}
  \end{minipage}
  \caption{Query planning on four multi-hop benchmarks. Lower copy rate and fewer calls indicate less broad or costly search; higher atomic-query rate and Correct@\(\leq3\) indicate more focused and effective retrieval.}
  \label{fig:query-planning-efficiency}
\end{figure}

\begin{figure}[t]
  \centering
  \includegraphics[width=\textwidth]{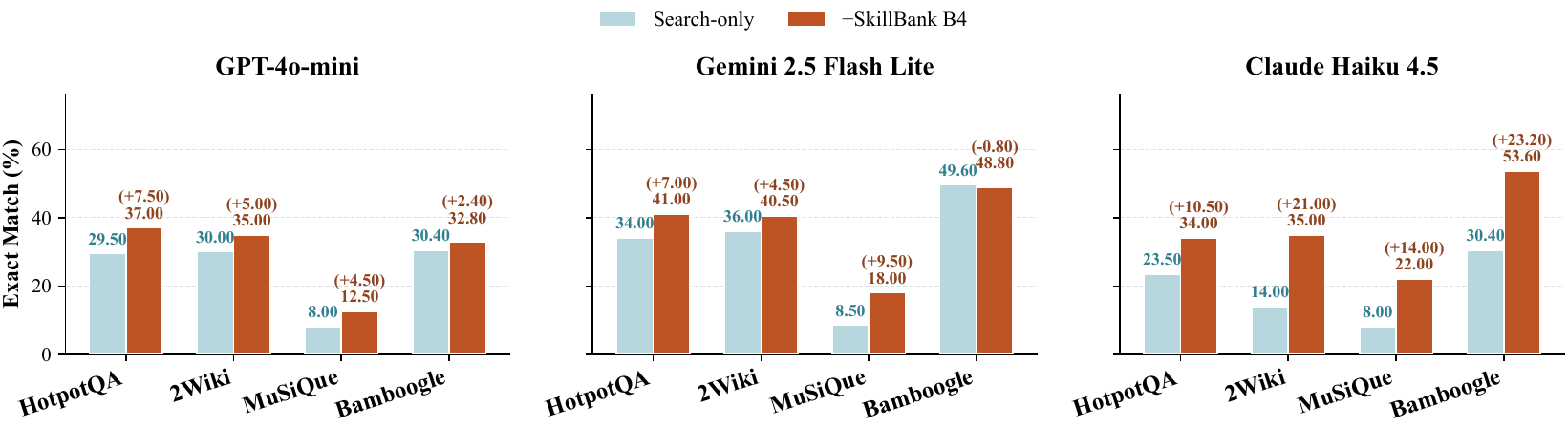}
  \caption{Closed-source transfer with frozen \(B_4\). Bars show EM percentages; parentheses give percentage-point gains over search-only prompting.}
  \label{fig:closed-source-skillbank}
\end{figure}

\subsection{Closed-source model transfer}
\label{sec:further-analysis-closed-source}

\begingroup
\looseness=-1
We also test whether the frozen SkillBank can serve as an external search prior for proprietary models without fine-tuning. Figure~\ref{fig:closed-source-skillbank} compares a search-only workflow with the same workflow augmented by \(B_4\) as an optional planning guide. The consistent macro gains across all three closed-source models show that learned search skills transfer beyond the trained open-source policies, and that exposing a structured SkillBank before retrieval can improve closed-source search planning.
\par\endgroup

\begingroup
\setlength{\intextsep}{4pt}
\FloatBarrier
\suppressfloats[t]
\begin{table}[t]
  \centering
  \setlength{\abovecaptionskip}{2pt}
  \setlength{\belowcaptionskip}{2pt}
  \caption{SkillBank construction (EM, \%). Avg. averages all seven benchmarks.}
  \label{tab:skillbank-construction}
  \small
  \setlength{\tabcolsep}{3pt}
  \renewcommand{\arraystretch}{1.0}
  \setlength{\aboverulesep}{1pt}
  \setlength{\belowrulesep}{2pt}
  \resizebox{\linewidth}{!}{%
  \begin{tabular}{lrrrrrrrr}
    \toprule
    Construction & HotpotQA & 2Wiki & MuSiQue & NQ & TriviaQA & PopQA & Bamboo. & Avg. \\
    \midrule
    Staged SkillBank & 41.67 & 47.92 & 13.89 & 31.25 & 47.92 & 43.60 & 28.80 & \textbf{36.44} \\
    One-shot SkillBank & 37.50 & 35.42 & 11.11 & 27.08 & 46.88 & 39.00 & 23.20 & \textbf{31.46} \\
    \bottomrule
  \end{tabular}}
  \par
\end{table}
\endgroup

\subsection{Why build the SkillBank in stages?}
\label{sec:skillbank-construction-ablation}
Staging allows skills to be refined as new tasks are introduced. In the \emph{one-shot} baseline, the 7B model instead receives all training questions and labels together and builds the entire bank at once.

As shown in Table~\ref{tab:skillbank-construction}, staging improves all seven benchmarks under matched training and evaluation settings, raising average EM from 31.46\% to 36.44\%.

\subsection{Reinforcement learning}
\label{sec:further-analysis-rl}
We further apply a short GRPO refinement stage after two-stage SFT while keeping the SkillBank interface fixed. The main RL results are reported in Table~\ref{tab:sft-rl-comparison}. Additional reward dynamics are provided in Appendix~\ref{app:rl-analysis}; this experiment probes how GRPO behaves in the SkillBank select-read-act action space and whether reinforcement learning can further shape skill execution.

\begingroup
\looseness=-1
The diagnostic bars in Figure~\ref{fig:rl-execution-diagnostics} show that, on SFT failures corrected by GRPO, RL conditions later actions on retrieved evidence more often (\(\uparrow\)), hits answer-containing evidence more frequently (\(\uparrow\)), grounds final answers more reliably (\(\uparrow\)), and reduces extra searches after evidence is found (\(\downarrow\)). These metrics suggest that GRPO mainly improves evidence use, answer grounding, and stopping behavior under the fixed SkillBank interface.
\par\endgroup

\begingroup
\setlength{\intextsep}{6pt}
\setlength{\abovecaptionskip}{6pt}
\setlength{\belowcaptionskip}{4pt}
\begin{table}[t]
  \centering
  \caption{RL results (EM fractions; Macro: unweighted mean). Parentheses give wrong-to-correct (green, +) and correct-to-wrong (red, --) changes from SFT as percentages of all questions.}
  \label{tab:sft-rl-comparison}
  \begingroup
  \setlength{\tabcolsep}{3pt}
  \renewcommand{\arraystretch}{1.05}
  \scriptsize
  \resizebox{\linewidth}{!}{%
  \begin{tabular}{lccccc}
    \toprule
    \textbf{Model} & \textbf{HotpotQA} & \textbf{2Wiki} & \textbf{MuSiQue} & \textbf{Bamboogle} & \textbf{Macro} \\
    \midrule
    Qwen2.5-7B-Instruct (GRPO) & \flipcell{0.440}{5.0}{4.0} & \flipcell{0.455}{11.0}{10.0} & \flipcell{0.145}{5.0}{6.0} & \flipcell{0.328}{4.8}{10.4} & \flipcell{0.342}{6.5}{7.6} \\
    Qwen2.5-7B-Base (GRPO) & \flipcell{0.400}{6.5}{8.5} & \flipcell{0.475}{7.5}{7.0} & \flipcell{0.100}{3.5}{8.0} & \flipcell{0.336}{4.0}{9.6} & \flipcell{0.328}{5.4}{8.3} \\
    Qwen2.5-3B-Instruct (GRPO) & \flipcell{0.355}{1.5}{1.5} & \flipcell{0.415}{2.0}{2.0} & \flipcell{0.160}{1.5}{1.5} & \flipcell{0.280}{4.0}{1.6} & \flipcell{0.303}{2.3}{1.7} \\
    Qwen2.5-3B-Base (GRPO) & \flipcell{0.365}{8.0}{5.5} & \flipcell{0.390}{12.5}{11.0} & \flipcell{0.135}{3.5}{3.5} & \flipcell{0.256}{4.8}{7.2} & \flipcell{0.287}{7.2}{6.8} \\
    \bottomrule
  \end{tabular}
  }
  \par
  \endgroup
  \vspace{4pt}
  \centering
  \includegraphics[width=0.60\linewidth]{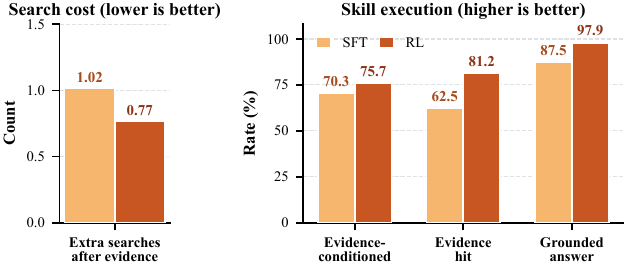}
  \captionof{figure}{Paired execution diagnostics for 7B-Instruct questions corrected by GRPO. Rates measure evidence use; extra searches count calls after answer-containing evidence is first retrieved.}
  \label{fig:rl-execution-diagnostics}
\end{table}
\endgroup


\par
\FloatBarrier
\section{Conclusion}
\label{sec:limitation-conclusion}
\label{sec:limitations-impact}
We present \Ours, a skill-conditioned framework that makes search-tool use explicit: the system evolves a reusable SkillBank, constructs bank-guided trajectories, and trains models to select skills before issuing retrieval or answer actions. Across open-source and closed-source models, our results show that explicit skill routing improves query planning, evidence use, and multi-hop answer grounding. The 7B model can also evolve new search skills on its own, eliciting its latent capabilities to achieve performance comparable to agents trained with large-scale reinforcement learning. Our auxiliary GRPO study does not yield large additional accuracy gains, but it noticeably changes execution behavior in the SkillBank action space, indicating that reinforcement learning can reshape how SFT policies select skills, search, and stop. Stronger RL algorithms and task-aligned rewards may further improve the framework.
\par

\FloatBarrier
\clearpage
\subsection*{AI use statement}
We used generative AI tools to help identify relevant literature, improve the manuscript's language, and debug and optimize a small portion of the code. Language models were also used to generate training trajectories as part of the proposed method, as detailed in Section~\ref{sec:searchskill}. The authors take full responsibility for the final manuscript, code, and reported results.

\begin{samepage}
\subsection*{Reproducibility statement}
All code and configurations required for training are available in the anonymous repository linked in the abstract. Section~\ref{sec:searchskill} specifies the framework and training objectives. Appendix~\ref{app:experimental-setups} provides data processing, skill cards, rollout pseudocode, and optimization settings, and Appendix~\ref{app:self-generation} documents the Qwen-generated implementation. Additional RL results, compute and API accounting, and case studies appear in Appendices~\ref{app:rl-analysis}, \ref{app:cost-analysis}, and~\ref{app:case-studies}. Data will be released through the same repository.
\par
\end{samepage}


\FloatBarrier
\bibliography{references}
\bibliographystyle{iclr2027_conference}

\clearpage
\appendix
\section{Related work}\label{app:related-work}
\subsection{RAG and RL}
Retrieval-augmented generation for knowledge-intensive QA began with retrieval-aware pretraining and dense retrieval backbones such as REALM and DPR~\citep{guu2020realm,karpukhin2020dpr}. RAG and FiD then made retrieved passages central to generation and reading~\citep{lewis2021rag,izacard2021fid}. Later systems such as RETRO, Atlas, and IRCoT further tightened the coupling between retrieval and reasoning~\citep{borgeaud2022retro,izacard2022atlas,trivedi2023ircot}. Dynamic retrieval policies were subsequently explored by Active RAG, Self-RAG, and unified active retrieval~\citep{jiang2023active,asai2023selfrag,cheng2024uar}.

In parallel, search became an explicit action space rather than a passive back-end. WebGPT and ReAct framed browsing and search as sequential reasoning-and-acting~\citep{nakano2022webgpt,yao2023react}, while Toolformer and Tool Learning generalized this pattern to broader tool use~\citep{schick2023toolformer,qin2024toollearning}. Search-o1 further embeds agentic search into long reasoning traces without reducing the problem to a static retrieve-then-read pipeline~\citep{li2025searcho1}. RL-based search agents then push this line further: R1-Searcher, DeepRetrieval, and Search-R1 directly optimize search behavior with reinforcement learning~\citep{song2025r1searcher,jiang2025deepretrieval,jin2025searchr1}. ZeroSearch complements them by improving search capability without live search during training~\citep{sun2025zerosearch}. Together, these works frame search as a trainable decision process rather than a fixed retrieval module.

\subsection{Skills}
Agent skills package instructions, resources, and executable code into reusable bundles that an agent can discover and load when relevant. A recent data-driven analysis of Claude Skills shows that such skills are already functioning as a practical extension layer for LLM agents~\citep{ling2026agentskillsdatadrivenanalysis}. At the same time, Agent Skills in the Wild highlights that once skills become reusable infrastructure, their organization and controlled invocation matter at scale~\citep{liu2026agentskillswildempirical}.

Recent research extends this idea from hand-authored skills to learned and evolving skill ecosystems. SkillWeaver shows that web agents can discover and hone reusable skills from interaction traces~\citep{zheng2025skillweaver}. Reinforcement Learning for Self-Improving Agent with Skill Library studies iterative policy improvement with an explicit skill library~\citep{wang2026skilllibraryrl}. CASCADE studies cumulative agentic skill creation through autonomous development and evolution, while AgentSkillOS investigates organization and orchestration at ecosystem scale~\citep{huang2026cascade,li2026agentskillos}. MemSkill reframes memory operations as evolvable skills~\citep{zhang2026memskill}, and SkillRL co-evolves a hierarchical SkillBank with the agent policy under reinforcement learning~\citep{xia2026skillrl}. Collectively, these works shift skills from hand-authored helpers to dynamic, reusable, and trainable components for agents, while study of skills for search-tool interaction remains limited.
\space 
\section{Self-generated skills and trajectories}
\label{app:self-generation}

The Qwen-generated setting uses the target Qwen2.5 backbone as \(\mathcal G\) in Section~\ref{sec:searchskill}. This appendix details the 7B run: Qwen2.5-7B-Instruct groups training questions, proposes and revises cards, executes search trajectories, and produces final answers. Bank evolution and trajectory generation use the fixed backbone; two-stage SFT then trains the search policy with the resulting bank and trajectories.

\subsection{Data grouping and bank evolution}
Training questions come from NQ, TriviaQA, HotpotQA, 2WikiMultiHopQA, and MuSiQue. Dataset-native metadata supplies coverage groups for multi-hop data. For NQ and TriviaQA, mean-pooled hidden states from the same 7B backbone are grouped by cosine \(k\)-means into eight clusters per dataset. Sampling balances these groups, with the number of questions per dataset allocated in proportion to the square root of the available pool size. Normalized question overlap with the available evaluation files is removed during manifest construction.

Starting from the six-card seed bank, each round uses 120 representative training questions to propose a candidate bank. The generator tests it through retrieval rollouts and receives their aggregate outcomes and compact success and failure traces. Skills used in successful trajectories are protected; unsuccessful candidates can be revised, merged, or rejected. Each card specifies When (trigger), Procedure (actions), Checkpoints (resolved facts), Continue (next-step conditions), Avoid (failure patterns), and Finalize (answer conditions). The four rounds produce banks of 8, 11, 12, and 13 skills. Their active datasets follow the NQ/TriviaQA, HotpotQA, 2Wiki, and MuSiQue order, with additional training questions from the other datasets included in each round.

\subsection{Trajectory construction and SFT}
The final 13-card bank guides rollouts over 8,000 training questions. The generator sees the question, its own prior actions, and retrieved passages when choosing queries and final answers. Exact match against the training answers, at least one retrieval action, and legal skill identifiers determine acceptance. This yields 2,859 successful trajectories.

Each accepted trajectory contributes a full-trajectory record and a closure-focused record that supervises only the final answer. An additional 21 records repeat trajectories using empirically rare successful skills. The resulting data contains 5,675 training records and 64 validation records. Stage I learns skill-tagged search and answer actions. Stage II rewrites each action into a selection turn, a selected-card context turn, and an execution turn, removing the question-specific recommended-skill hint. The stages use the same learning rates, sequence lengths, loss weights, and LoRA configuration listed in Table~\ref{tab:appendix-two-stage-sft-config}.

\begin{table}[t]
\centering
\caption{Trajectory generation with the Qwen2.5-7B-Instruct generator.}
\label{tab:self-generation-data}
\small
\begin{tabular}{lrr}
\toprule
Dataset & Training questions & Accepted trajectories \\
\midrule
NQ & 1,021 & 318 \\
TriviaQA & 1,021 & 419 \\
HotpotQA & 2,247 & 1,063 \\
2WikiMultiHopQA & 1,835 & 838 \\
MuSiQue & 1,876 & 221 \\
\midrule
Total & 8,000 & 2,859 \\
\bottomrule
\end{tabular}
\end{table}

\subsection{Evaluation}
The seven-benchmark evaluation contains 200 questions each from HotpotQA, 2WikiMultiHopQA, and MuSiQue; 1,000 each from NQ, TriviaQA, and PopQA; and all 125 Bamboogle questions. The Qwen-generated 7B policy obtains 38.60 macro EM, and the reported Qwen-generated 3B policy obtains 34.87, as shown in Table~\ref{tab:main-results}.

\section{Reinforcement learning analysis}
\label{app:rl-analysis}

We further apply a short GRPO refinement stage after two-stage SFT while keeping the SkillBank interface fixed. As an auxiliary test of post-SFT optimization, we use a lightweight reward
\[
r = r_{\mathrm{EM}} + \lambda_e r_{\mathrm{evi}} - \lambda_d r_{\mathrm{dup}},
\]
where the terms reward final-answer correctness and answer-containing evidence while discouraging duplicate searches. The main RL results are reported in Table~\ref{tab:sft-rl-comparison}; for each run, we select the reported checkpoint according to training diagnostics, including reward trends, valid-search behavior, and development scores, rather than always using the final checkpoint. Additional reward dynamics are provided in Appendix~\ref{app:grpo-reward-dynamics}; this experiment probes how GRPO behaves in the SkillBank select-read-act action space and whether reinforcement learning can further shape skill execution.

\begin{figure}[t]
  \centering
  \includegraphics[width=\linewidth]{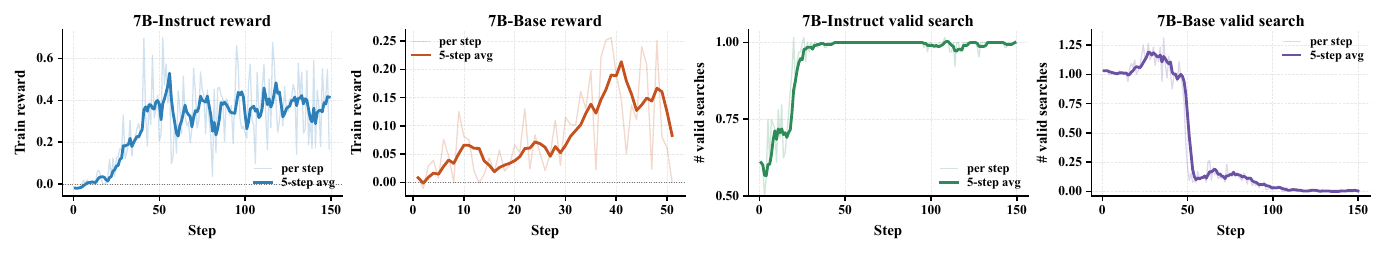}
  \caption{GRPO training diagnostics for 7B models. Panels show mean rollout reward and valid search calls per rollout. A valid call is a parsed search action executed by the retriever.}
  \label{fig:rl-train-reward}
\end{figure}

\noindent Figure~\ref{fig:rl-train-reward} reports reward and valid-search dynamics for the 7B GRPO runs. The 7B-Instruct policy shows a noisy but sustained increase in train reward while keeping valid search usage near one search per rollout. The 7B-Base run initially improves, but its valid-search count later collapses, so we truncate its reward panel at the collapse point and treat this behavior as a reward-design limitation rather than a stable gain. This contrast motivates using both reward and action-validity diagnostics when selecting RL checkpoints.

Across all 725 paired 7B-Instruct evaluation questions, mean search calls decrease from 2.98 to 2.48 after GRPO. The mean number of calls after the first answer-containing passage falls from 0.60 to 0.42. Both means include every question; a trajectory that never retrieves answer-containing evidence contributes zero to the latter metric. Macro EM across the four datasets changes from 35.35\% to 34.20\%.

Figure~\ref{fig:rl-execution-diagnostics} compares the SFT and RL trajectories on questions that change from wrong to correct. \emph{Evidence-conditioned} tracks subsequent queries that use information from earlier retrieval. \emph{Evidence hit} records retrieval of an answer-containing passage, and \emph{Grounded answer} records agreement between the final answer and its supporting evidence. \emph{Extra searches after evidence} counts retrieval calls after the first answer-containing passage is found. The first three are rates (\(\uparrow\)); the last is a mean count (\(\downarrow\)). On this corrected subset, GRPO improves evidence use and grounding while reducing additional searches.
\FloatBarrier

\section{Experimental details}
\label{app:experimental-setups}

SearchSkill uses the same stage-wise bank evolution, trajectory construction, and two-stage SFT framework across generator choices. The detailed data-processing recipe and skill taxonomy below describe the GPT-5.4-generated instantiation; Appendix~\ref{app:self-generation} records the Qwen-generated implementation.

\subsection{Data process}
\label{app:data-process}

We construct the training pool with a coverage-oriented data processing pipeline rather than uniform random sampling. The pipeline first reads the cached training splits and removes direct overlap with the corresponding evaluation ids when such ids are available. It then converts each example into a lightweight profile containing the question, answer form, question type, coarse entity count, length bucket, dataset-native type or hop count, and heuristic retrieval cues such as temporal, numerical, comparison, alias, verification, and relation-chain indicators. These profiles are grouped into coarse reasoning signatures, which serve as the sampling units for preserving diversity.

For single-hop datasets, Natural Questions and TriviaQA are sampled with a signature-capped strategy. Rare signatures are protected, frequent signatures are capped, and the remaining examples are ranked to improve label coverage over answer forms, WH types, question length, entity density, and retrieval-related cues. This produces 3,000 examples from Natural Questions and 3,000 examples from TriviaQA. For multi-hop datasets, HotpotQA, 2WikiMultiHopQA, and MuSiQue are first grouped by signatures that include native question type and hop count. We then ask \textsc{GPT-5.4} to annotate representative groups with reasoning types, skill demands, answer form, entity pattern, difficulty, and coverage priority. The sampler uses these annotations to enforce coverage over bridge reasoning, comparison, relation chains, temporal and numerical reasoning, set constraints, verification, and answer composition. After initial selection, the pipeline removes duplicate or conflicting questions, reviews suspicious records with the same model, and refills dropped items from the ranked candidate pool.

\begin{table}[tbp]
  \centering
  \caption{Coverage-oriented training subsets produced by the data processing pipeline.}
  \label{tab:appendix-data-process}
  \setlength{\tabcolsep}{7pt}
  \renewcommand{\arraystretch}{1.1}
  \small
  \begin{tabular}{lrrr}
    \toprule
    \textbf{Dataset} & \textbf{Original examples} & \textbf{Reasoning signatures} & \textbf{Selected examples} \\
    \midrule
    Natural Questions & 79,168 & 1,086 & 3,000 \\
    TriviaQA & 78,785 & 5,130 & 3,000 \\
    HotpotQA & 90,447 & 3,094 & 12,445 \\
    2WikiMultiHopQA & 15,000 & 410 & 8,300 \\
    MuSiQue & 19,938 & 2,118 & 8,594 \\
    \bottomrule
  \end{tabular}
\end{table}

The output of this stage consists of lightweight profiles, group annotations, sampling reports, and final full-record JSONL files. The sampled examples are then passed to the trajectory construction pipeline, where teacher rollouts, retriever interactions, and positive-trajectory filtering produce the SFT supervision described in Section~\ref{sec:searchskill}.

\subsection{SkillBank evolution}
\label{app:skillbank-evolution}

The SkillBank is evolved before trajectory SFT because the search interface should expose reusable retrieval policies rather than dataset-specific answer shortcuts. If the bank is too small, the model must express distinct behaviors with the same generic search action; if it is expanded without structure, the model faces a noisy routing problem. We therefore grow the bank in curriculum order, starting from a compact \(B_0\) that contains the core bridge, comparison, temporal, conflict-checking, evidence-span, and answer-grounding skills, and then adding only skills that correspond to stable behavior patterns observed in the sampled data.

Each evolution round uses the processed samples from Appendix~\ref{app:data-process}. The examples are grouped into discovery packets by reasoning signature, and each packet keeps a small set of representative questions together with metadata such as dataset, group size, native type, hop count, answer form, entity pattern, and heuristic flags. These packets are summarized into coverage buckets, then \textsc{GPT-5.4} proposes a new SkillBank version from the previous bank, the bucket summaries, and representative examples. The update is constrained to preserve existing useful skills, add skills only when they change search planning or verification behavior, and refine old skills when their routing boundary is too broad. This design makes the bank an explicit external prior for later teacher synthesis, rather than something the student must rediscover from SFT labels alone.

\begin{table}[tbp]
  \centering
  \caption{Curriculum used to evolve the SearchSkill SkillBank. New skills are added only when the corresponding data stage exposes a reusable search or verification behavior.}
  \label{tab:appendix-skillbank-evolution}
  \setlength{\tabcolsep}{4pt}
  \renewcommand{\arraystretch}{1.12}
  \small
  \begin{tabular}{llcp{0.48\linewidth}}
    \toprule
    \textbf{Bank} & \textbf{Evolution data} & \textbf{Added} & \textbf{Main purpose of the update} \\
    \midrule
    \(B_0\) & Seed bank & 6 & Core bridge search, comparison, temporal extraction, conflict checking, evidence extraction, and grounding. \\
    \(B_1\) & NQ + TriviaQA & 5 & Add a single-hop spine for direct relation lookup, surface-name resolution, anchored queries, ranking lookup, and forced-choice answers. \\
    \(B_2\) & HotpotQA & 3 & Add reusable two-hop bridge planning, bridge disambiguation, relation-chain decomposition, and multihop yes/no verification. \\
    \(B_3\) & 2WikiMultiHopQA & 3 & Add bridge-comparison planning, derived-kinship inference joins, and checkpointed longer-hop retrieval. \\
    \(B_4\) & MuSiQue & 3 & Add long-hop re-anchoring, temporal-anchor carry-forward, and full-chain reconstruction before final answering. \\
    \bottomrule
  \end{tabular}
\end{table}

This staged evolution is important for two reasons. First, it separates skill discovery from policy optimization: the bank is built by analyzing broad coverage packets, while the student later learns to invoke the resulting skills through explicit tags. Second, it avoids asking the model to choose from a large long-hop bank before it has stable simpler skills. Direct lookup and answer grounding are established first, bridge and comparison skills are added next, and only then do we introduce long-hop decomposition and recovery skills for MuSiQue-style questions. The final \(B_4\) is therefore a cumulative skill prior that covers single-hop lookup, two-hop bridge reasoning, compositional comparison, and longer decomposition under one consistent interface.

\begin{table}[tbp]
  \centering
  \caption{GPT-5.4-generated \(B_4\), grouped by the four functional categories in Section~\ref{sec:further-analysis-skillbank-evolution}. Summaries are condensed from the cards; the Qwen2.5-7B-generated bank is described in Appendix~\ref{app:self-generation}.}
  \label{tab:appendix-b4-skillbank-categories}
  \setlength{\tabcolsep}{3pt}
  \renewcommand{\arraystretch}{1.02}
  \scriptsize
  \begin{tabular}{@{}p{0.18\linewidth}p{0.36\linewidth}p{0.40\linewidth}@{}}
  \toprule
  \textbf{Category} & \textbf{Skill} & \textbf{Skill-card summary} \\
  \midrule
  Direct lookup &
  \nolinkurl{single-entity-relation-lookup} &
  Query one resolved entity plus the requested relation or attribute, then verify that the evidence explicitly attaches the answer to that entity. \\
  &
  \nolinkurl{surface-name-resolution} &
  Resolve real names, alternate names, former names, nicknames, or character-to-actor mappings from contextual name-linking evidence. \\
  &
  \nolinkurl{superlative-ranking-match} &
  Retrieve and verify a superlative claim such as first, largest, highest, oldest, or top-ranked within the specified set and timeframe. \\
  &
  \nolinkurl{forced-choice-option-resolution} &
  Treat an explicit option set in the question as answer anchors, verify which option matches the evidence, and answer with that option span. \\
  &
  \nolinkurl{temporal-range-extract} &
  Extract a date, year, count, measurement, rank, or range for an already identified target while preserving unit, scope, and timeframe. \\
  \midrule
  Bridge/chain &
  \nolinkurl{bridge-entity-search} &
  Find one hidden intermediate entity from a named clue, then search the final attribute on that confirmed bridge entity. \\
  &
  \nolinkurl{bridge-disambiguate-then-hop} &
  Resolve an alias-heavy or underspecified bridge candidate with rare anchors before launching the second-hop attribute search. \\
  &
  \nolinkurl{relation-chain-decomposition} &
  Follow an explicit ordered relation chain one edge at a time, verifying each intermediate before continuing to the next relation. \\
  &
  \nolinkurl{multi-constraint-query-anchoring} &
  Build a query from several rare constraints when the next target is clue-heavy or alias-prone, then re-anchor after the candidate is verified. \\
  &
  \nolinkurl{sequential-hop-checkpointing} &
  Preserve multiple intermediate checkpoints in 3-plus-hop or mirrored multi-step questions, restarting from the last secure checkpoint when needed. \\
  &
  \nolinkurl{re-anchored-long-hop-decomposition} &
  Solve long clue-heavy paths by making each confirmed entity the new anchor for the unresolved remainder of the question. \\
  &
  \nolinkurl{temporal-anchor-carry-forward} &
  Attach a temporal qualifier to the hop it constrains, carry the time-scoped checkpoint forward, and verify downstream evidence under that timeframe. \\
  \midrule
  Comparison/join &
  \nolinkurl{parallel-attribute-compare} &
  Retrieve matched evidence for two named entities under the same attribute, unit, definition, and timeframe before comparing or selecting. \\
  &
  \nolinkurl{bridge-comparison-planning} &
  Resolve hidden bridge entities on both sides of a comparison before retrieving endpoint attributes and making the final comparison. \\
  &
  \nolinkurl{derived-kinship-inference-join} &
  Derive family roles such as paternal grandmother or in-law relations by joining immediate relative and second-edge family evidence. \\
  &
  \nolinkurl{multihop-yes-no-verification} &
  Decompose a yes/no claim into required subclaims, gather matched evidence for each, and answer yes only when all are explicitly supported. \\
  \midrule
  Grounding/verification &
  \nolinkurl{conflict-check} &
  Run a targeted disambiguation search when retrieved evidence surfaces competing entities, titles, dates, numbers, or inconsistent checkpoints. \\
  &
  \nolinkurl{verbatim-evidence-span} &
  Copy the shortest faithful answer span directly supported by evidence, preserving names, titles, units, quotations, or list structure. \\
  &
  \nolinkurl{answer-grounding-check} &
  Apply a final grounding checklist over answer support, relation match, answer type, units, timeframe, and all bridge or comparison edges. \\
  &
  \nolinkurl{reconstructed-chain-verification} &
  Replay confirmed checkpoints before answering long chains, ensuring the endpoint answer belongs to the intended final target rather than a nearby entity. \\
  \bottomrule
  \end{tabular}
\end{table}

\subsection{Trajectory construction}
\label{app:trajectory-construction}

Trajectory construction turns the sampled examples and the evolved SkillBank into executable supervision. We first build a manifest from the processed training pools in Appendix~\ref{app:data-process}. Each manifest item stores the question, gold answers, dataset family, metadata summary, candidate primary skills, and suggested support skills. The manifest also mixes a small number of failure-replay examples from earlier evaluations, so the teacher can reconstruct trajectories for cases where previous policies selected poor skills, stopped too early, or answered with an intermediate entity.

For each manifest item, \textsc{GPT-5.4} acts as a teacher policy under the current SkillBank. At every step, the teacher returns a structured JSON action containing a primary skill, optional support skills, an action type, a search query or draft answer, and a checkpoint of the current resolved state. If the action is \texttt{search} or \texttt{verify}, the local retriever is called with top-\(3\) retrieval and the returned passages are appended to the trace. If the action is \texttt{answer}, a finalizer prompt normalizes the draft into the shortest answer span that is explicitly supported by the collected evidence, typically using \texttt{verbatim-evidence-span} together with \texttt{answer-grounding-check}. This produces raw trajectories with interleaved skills, actions, retrieved evidence, checkpoints, and final answers.

\begin{algorithm}[H]
  \caption{Two-phase SkillBank interface in SearchSkill}
  \label{alg:searchskill-rollout}
  \begin{algorithmic}[1]
    \Require Question \(x\), policy model, SkillBank \(B^\star\), retriever, budget \(T\)
    \Ensure Final answer and skill-annotated interaction trace
    \State Build compact index \(I=\mathrm{Index}(B^\star)\)
    \State Initialize evidence \(E_0\leftarrow\emptyset\), prompt state \(H_0\leftarrow[x,I]\), trace \(\Gamma\leftarrow\emptyset\)
    \For{\(t=1,\ldots,T\)}
      \State Generate skill-selection turn and parse \(s_t\) from \tagopen{selecttagcolor}{select\_skill}...\tagclose{selecttagcolor}{select\_skill}
      \State Load selected cards \(C_t\leftarrow\mathrm{Cards}(B^\star,s_t)\)
      \State Generate action turn \tagopen{skilltagcolor}{skill}\(\hat{s}_t\)\tagclose{skilltagcolor}{skill} followed by one action tag
      \State Parse \(o_t\in\{\tagopen{searchtagcolor}{search},\tagopen{answertagcolor}{answer}\}\) and content \(z_t\)
      \State Update trace \(\Gamma\leftarrow\Gamma\cup\{(s_t,C_t,o_t,z_t)\}\)
      \If{\(o_t=\textsc{Search}\)}
        \State Execute retriever \(R_t\leftarrow\mathrm{Search}(z_t)\)
        \State Update evidence \(E_{t+1}\leftarrow E_t\cup R_t\)
        \State Update prompt \(H_{t+1}\leftarrow H_t+(s_t,C_t,o_t,z_t)+\tagopen{infotagcolor}{information}R_t\tagclose{infotagcolor}{information}+I\)
      \ElsIf{\(o_t=\textsc{Answer}\)}
        \State \Return final answer \(z_t\) and trace \(\Gamma\)
      \Else
        \State Stop with invalid action
      \EndIf
    \EndFor
    \State Stop when the search budget is exhausted
  \end{algorithmic}
\end{algorithm}

We keep only trajectories that pass executable validation. A trajectory must answer correctly under exact-match normalization, contain at least one retrieval step, use only legal SkillBank identifiers, and avoid using support-only verification skills as the primary skill before the final answer. When multiple rollout sources contain the same example, we keep the higher-quality row according to validation success, runtime status, nonempty final answer, legal skill usage, route consistency, and trace completeness. The early canonical teacher set contains 360 validated trajectories from 757 unique candidates, but the final SFT pool is larger because we continue with coverage-supplement rollouts and replay earlier validated data.

\begin{table}[tbp]
  \centering
  \caption{Final trajectory-format SFT pool. The train split reuses validated replay trajectories and adds cleaned coverage-supplement trajectories.}
  \label{tab:appendix-trajectory-sft-pool}
  \setlength{\tabcolsep}{5pt}
  \renewcommand{\arraystretch}{1.1}
  \small
  \begin{tabular}{lrp{0.48\linewidth}}
    \toprule
    \textbf{Source} & \textbf{Records} & \textbf{Role in the SFT pool} \\
    \midrule
    Replay train records & 1,445 & Preserve earlier validated SearchSkill supervision \\
    Cleaned supplement trajectories & 786 & Add broader coverage after filtering and deduplication \\
    Closure-expanded supplement records & 1,572 & Add action and closure supervision from the supplement \\
    Final train split & 3,017 & Trajectory-format SFT training data \\
    Final eval split & 16 & Held-out SFT validation records \\
    \bottomrule
  \end{tabular}
\end{table}

The final SFT data is therefore not limited to the 360 early canonical trajectories. The trajectory-format training split contains 3,017 records: 1,445 records from the earlier validated replay set and 1,572 additional records derived from 786 cleaned coverage-supplement trajectories after closure expansion. These records supervise the model to emit explicit \tagopen{skilltagcolor}{skill}...\tagclose{skilltagcolor}{skill} declarations before each \tagopen{searchtagcolor}{search}...\tagclose{searchtagcolor}{search} or \tagopen{answertagcolor}{answer}...\tagclose{answertagcolor}{answer} action, while the held-out SFT validation split contains 16 records.

\begin{table}[tbp]
  \centering
  \caption{Question-type distribution of the trajectory-format SFT training data. ``Unknown'' mainly corresponds to the single-hop NQ and TriviaQA records, where no dataset-native multihop type is defined.}
  \label{tab:appendix-sft-type-distribution}
  \setlength{\tabcolsep}{6pt}
  \renewcommand{\arraystretch}{1.1}
  \small
  \begin{tabular}{lrr}
    \toprule
    \textbf{Question type} & \textbf{Records} & \textbf{Main source} \\
    \midrule
    Bridge & 682 & HotpotQA \\
    Comparison & 561 & HotpotQA / 2WikiMultiHopQA \\
    Compositional & 371 & 2WikiMultiHopQA \\
    MuSiQue decomposition-2 & 201 & MuSiQue \\
    MuSiQue decomposition-3 & 217 & MuSiQue \\
    MuSiQue decomposition-4 & 156 & MuSiQue \\
    Bridge comparison & 104 & 2WikiMultiHopQA \\
    Inference & 95 & 2WikiMultiHopQA \\
    Single-hop / unlabeled & 630 & NQ / TriviaQA / residual records \\
    \bottomrule
  \end{tabular}
\end{table}

\subsection{Two-stage SFT training}
\label{app:two-stage-sft-training}

This section reports the 7B-Instruct SFT configuration used for the main SearchSkill policy. Both stages use Qwen2.5-7B-Instruct as the backbone and train LoRA adapters with the same target modules: \texttt{q\_proj}, \texttt{k\_proj}, \texttt{v\_proj}, \texttt{o\_proj}, \texttt{up\_proj}, \texttt{down\_proj}, and \texttt{gate\_proj}. The first stage trains the execution format from the validated trajectory-format supervision. The second stage initializes from the first-stage adapter and trains the SkillBank-card protocol, where each original action turn is rewritten into a skill-selection turn, a selected-card context turn, and the original skill-grounded action turn.

\begin{table}[tbp]
  \centering
  \caption{Two-stage SFT configuration for the Qwen2.5-7B-Instruct SearchSkill policy.}
  \label{tab:appendix-two-stage-sft-config}
  \setlength{\tabcolsep}{4pt}
  \renewcommand{\arraystretch}{1.12}
  \small
  \begin{tabular}{p{0.25\linewidth}p{0.31\linewidth}p{0.31\linewidth}}
    \toprule
    \textbf{Item} & \textbf{Stage I: execution SFT} & \textbf{Stage II: SkillBank-card SFT} \\
    \midrule
    Initialization & Qwen2.5-7B-Instruct & Stage-I adapter on Qwen2.5-7B-Instruct \\
    Supervision format & \tagopen{skilltagcolor}{skill} plus one \tagopen{searchtagcolor}{search} or \tagopen{answertagcolor}{answer} action & \tagopen{selecttagcolor}{select\_skill}, selected skill cards, then \tagopen{skilltagcolor}{skill} plus one action \\
    Train / eval records & 3,017 / 16 & 3,017 / 16 \\
    Max sequence length & 8,192 & 12,288 \\
    Truncation side & Left & Left \\
    Learning rate & \(7\times10^{-5}\) & \(1\times10^{-5}\) \\
    Epochs & 2.0 & 1.0 \\
    Warmup ratio & 0.05 & 0.03 \\
    GPUs & 4 & 4 \\
    Per-device train batch & 1 & 1 \\
    Gradient accumulation & 4 & 4 \\
    Effective train batch & 16 & 16 \\
    LoRA setting & \(r=16\), \(\alpha=32\), dropout \(0.05\) & \(r=16\), \(\alpha=32\), dropout \(0.05\) \\
    Loss weights & answer \(2.5\), search \(0.8\), other \(1.0\) & answer \(2.0\), search \(0.8\), other \(1.0\) \\
    Scheduler / precision & cosine; bf16 when supported, otherwise fp16 & cosine; bf16 when supported, otherwise fp16 \\
    Logging / eval / save steps & 5 / 40 / 40 & 5 / 40 / 40 \\
    \bottomrule
  \end{tabular}
\end{table}

During preprocessing, user and system tokens are masked out and only assistant outputs are supervised. For records marked as full-trajectory supervision, every assistant turn contributes to the loss; for closure-focused records, only the final assistant turn is supervised. The weighted token loss assigns larger weight to answer turns to improve grounded stopping, while search turns receive a smaller weight to avoid over-optimizing surface query strings.

\begin{figure}[tbp]
  \centering
  \includegraphics[width=\linewidth]{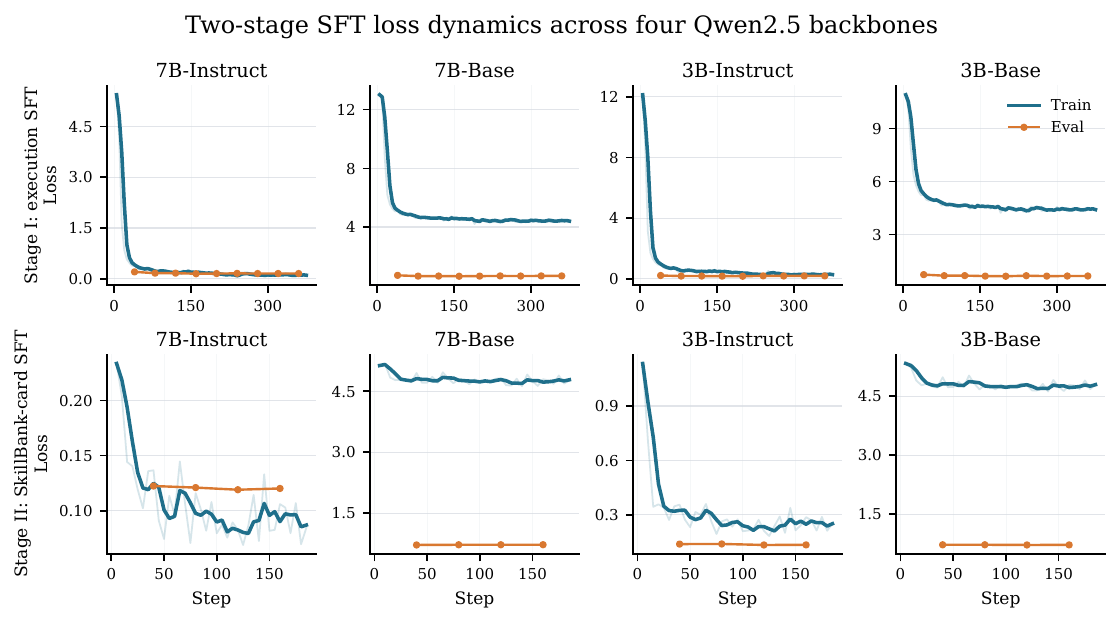}
  \caption{Training and validation loss curves for the two-stage SFT runs across four Qwen2.5 backbones. Each row corresponds to one SFT stage, and columns correspond to model variants. Thin lines show raw training loss and thick lines show a three-point moving average.}
  \label{fig:appendix-sft-two-stage-loss}
\end{figure}

\textbf{SFT loss dynamics.} Figure~\ref{fig:appendix-sft-two-stage-loss} summarizes the training curves for the two supervised stages. Across both stages, instruction-tuned backbones absorb the SearchSkill format more easily than base backbones: their validation losses are lower, and the Stage-II SkillBank-card objective remains stable after initializing from Stage I. Base models still reduce loss under the same supervision, but their much higher training loss suggests that learning the structured search protocol is harder without prior instruction-following behavior. This supports using the two-stage recipe as a portable training protocol while expecting stronger and cleaner convergence from instruction-tuned initializations.

\subsection{Definitions of diagnostic metrics}
\label{app:diagnostic-metrics}

\textbf{Skill activation and Rescue in Figure~\ref{fig:skillbank-evolution}.}
The evaluation set \(\mathcal D\) contains 725 questions: 200 each from HotpotQA, 2WikiMultiHopQA, and MuSiQue, and 125 from Bamboogle. For category \(c\), let \(A_c\) contain the questions whose trajectory uses at least one skill in that category. Each question is counted once per category, even if the category is used repeatedly. Let \(N_c\subseteq A_c\) contain the questions with an EM-correct answer for which the judge identifies a necessary contribution from that category. We report
\[
\operatorname{Activation}(c)=100\frac{|A_c|}{|\mathcal D|},
\qquad
\operatorname{Rescue}(c)=100\frac{|N_c|}{|A_c|}.
\]
Activation measures how often a category is used; Rescue measures how often its use is judged necessary in a successful answer. Incorrect activated examples remain in the denominator and contribute zero to the numerator.

Only correct activated traces are sent to GPT-4o-mini at temperature 0. The judge receives the question, reference and predicted answers, skill ids, and ordered actions with queries and retrieved evidence. It checks whether removing the category's steps would lose a required fact, intermediate entity, comparison decision, or evidence-based correction. A positive judgment must cite concrete step ids. A decisive lookup, required bridge, comparison, or correction qualifies; routine formatting and repeated information do not. Outputs contain \texttt{rescued}, \texttt{step\_ids}, and \texttt{reason}. Rescue is a judgment about the recorded trajectory. The category-removal experiment instead removes the corresponding cards from the available bank and reruns inference to measure the change in EM.

\textbf{Query planning in Figure~\ref{fig:query-planning-efficiency}.}
\emph{First query copy} measures the fraction of first queries that reproduce or paraphrase the question as a whole, retaining its unresolved constraints instead of choosing a retrieval target. \emph{Atomic-hop query} measures the fraction of all search calls directed at one retrieval subgoal: an entity, relation, attribute, bridge, comparison operand, or verification fact. For example, a birthplace query about an already resolved person is atomic; repeating a question that still requires resolving that person's identity is not. \emph{Average search} is the number of retrieval calls divided by the number of questions. \emph{Correct@\(\leq3\) searches} is the percentage of all questions for which EM is correct and the complete trajectory uses at most three retrieval calls. It includes both correctness and cost in its numerator; it is not accuracy conditioned on short trajectories. Each diagnostic is first computed within a dataset and then averaged equally across the four multi-hop datasets.


\subsection{GRPO reward dynamics}
\label{app:grpo-reward-dynamics}

Our RL action space follows the same fixed SkillBank interface used at inference time: the policy selects skill ids, reads the corresponding card content, and then emits either a search query or a final answer. We use this section only to report auxiliary reward dynamics rather than to introduce a separate RL method; the reward used for the post-SFT test is summarized in Section~\ref{sec:further-analysis-rl}.

\begin{table}[tbp]
  \centering
  \caption{GRPO configuration for the 7B diagnostic runs. Figure~\ref{fig:appendix-rl-7b-base-dynamics} reports the available cumulative 7B-Base training-reward curve.}
  \label{tab:appendix-grpo-config}
  \small
  \begin{tabular}{>{\raggedright\arraybackslash}p{0.25\linewidth}>{\raggedright\arraybackslash}p{0.32\linewidth}>{\raggedright\arraybackslash}p{0.32\linewidth}}
    \toprule
    Setting & 7B-Instruct & 7B-Base \\
    \midrule
    Initialization & Two-stage SFT policy & Two-stage SFT policy \\
    Training / validation examples & 1200 / 80 & 1200 / 80 \\
    GPUs / parallelism & 4 H20, FSDP offload & 4 H20, FSDP offload \\
    GRPO samples & 8 rollouts per prompt & 8 rollouts per prompt \\
    Train / validation batch & 8 / 80 & 8 / 80 \\
    PPO mini / micro batch & 32 / 4 & 32 / 4 \\
    Actor learning rate & \(1\times10^{-6}\) & \(1\times10^{-6}\) \\
    KL regularization & low-variance KL, coefficient \(0.001\) & low-variance KL, coefficient \(0.001\) \\
    Entropy coefficient / grad clip & \(0.001\) / \(1.0\) & \(0.001\) / \(1.0\) \\
    Rollout decoding & temperature \(1.1\), top-\(p=0.98\) & temperature \(1.1\), top-\(p=0.98\) \\
    Search interface & top-3 retriever, max 5 turns & top-3 retriever, max 5 turns \\
    Length limits & prompt 8192, response 192, observation 1400 & prompt 8192, response 192, observation 1400 \\
    Eval / save frequency & every 15 steps / every 15 steps & every 15 steps / every 15 steps \\
    Planned optimization steps & 150 & 150 \\
    \bottomrule
  \end{tabular}
\end{table}

\begin{figure}[tbp]
  \centering
  \includegraphics[width=\linewidth]{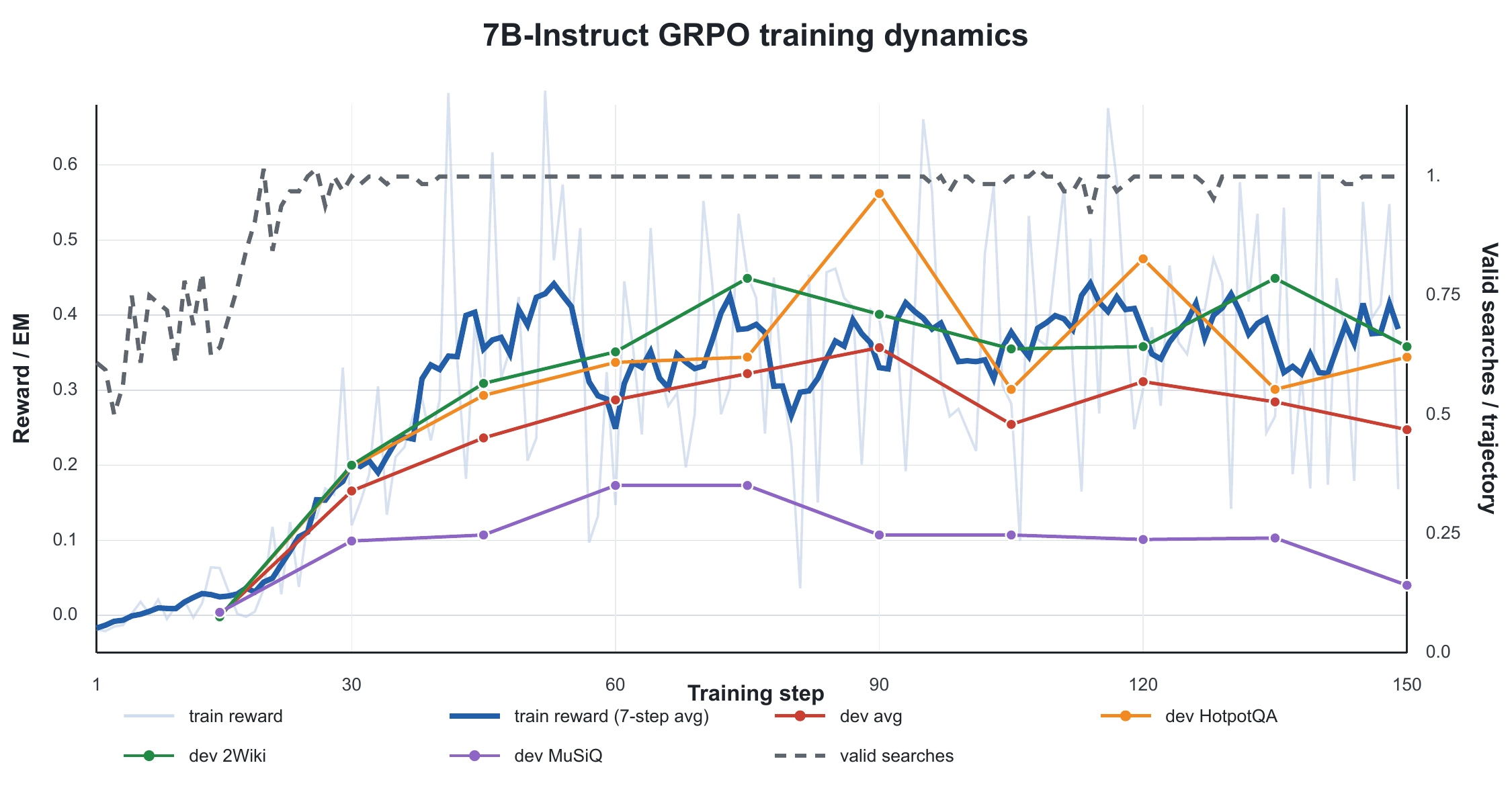}
  \caption{GRPO training dynamics for the 7B-Instruct SFT-initialized policy. The plot shows train reward, smoothed train reward, validation exact-match curves, and valid-search usage.}
  \label{fig:appendix-rl-process-reward}
\end{figure}

\begin{figure}[tbp]
  \centering
  \includegraphics[width=\linewidth]{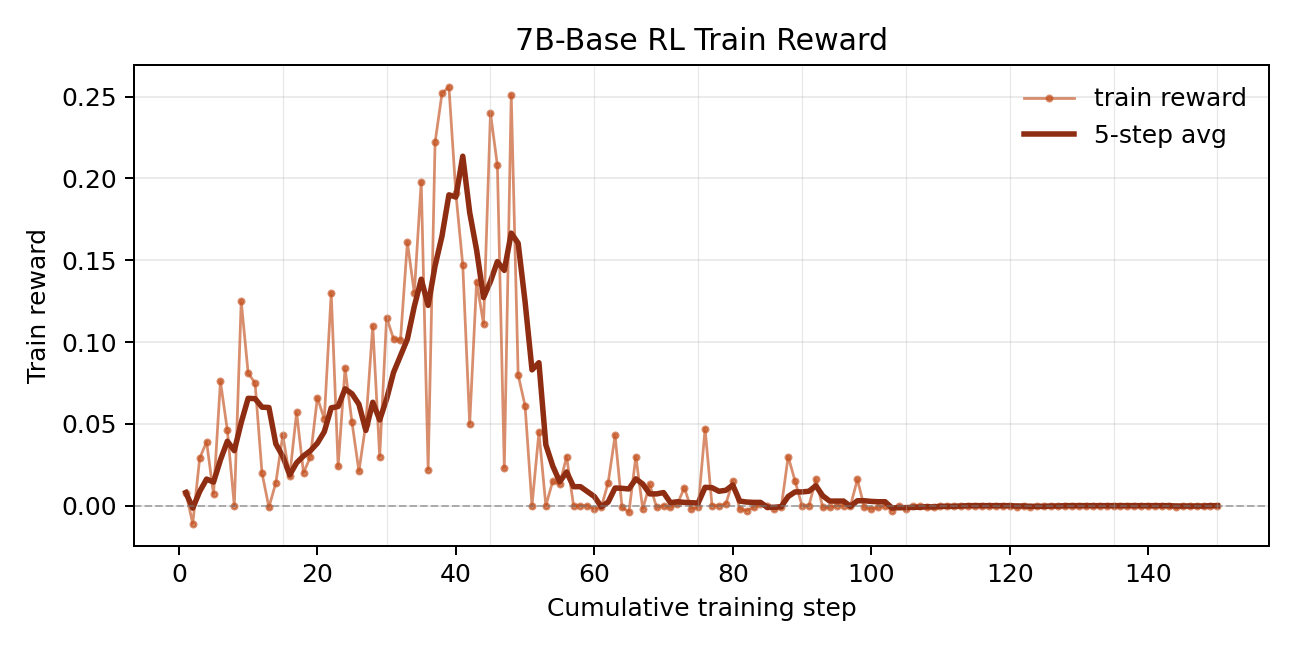}
  \caption{Cumulative GRPO training reward for the 7B-Base SFT-initialized policy. The plot shows raw train reward and a five-step moving average over cumulative training steps.}
  \label{fig:appendix-rl-7b-base-dynamics}
\end{figure}

Figure~\ref{fig:appendix-rl-process-reward} shows the 7B-Instruct GRPO trajectory initialized from two-stage SFT: training reward rises after early skill-space exploration while valid search usage remains stable, but the validation curves remain noisy across benchmarks. Figure~\ref{fig:appendix-rl-7b-base-dynamics} reports the cumulative 7B-Base training-reward curve; reward rises during the early and middle training phase, then becomes unstable and decays toward the end of the run. These diagnostics support treating RL as a promising but still exploratory extension for directly optimizing skill selection, search execution, and grounded stopping beyond SFT imitation.

\section{Limitations and future directions}
\label{app:limitations}

SearchSkill currently evolves its SkillBank for multi-hop question answering from a finite, coverage-oriented training pool. This makes the learned bank effective for the reasoning patterns seen in our data, but it may miss skills that only appear in new domains, new tools, or rare reasoning forms. Extending the framework therefore requires more complete mechanisms for continual skill discovery, validation, merging, and revision when the task distribution changes.

Our reinforcement-learning results also reveal both flexibility and risk. GRPO can substantially change the action habits of an SFT policy in the select-read-act space, but it can also disturb behavior on multi-hop questions that the SFT model already solves, for example by changing search frequency or weakening stable stopping behavior. Better algorithms and rewards that are explicitly matched to SkillBank routing, retrieval execution, and grounded answering are needed; with such alignment, reinforcement learning may further improve the capability of the overall workflow.

\section{Existing assets and licenses}
\label{app:asset-licenses}

SearchSkill uses publicly released benchmarks, model checkpoints, retrievers, and baseline implementations only for academic research. Tables~\ref{tab:appendix-benchmark-licenses} and~\ref{tab:appendix-model-baseline-licenses} record the main existing assets used or cited by the experimental setup; the corresponding source papers are cited in the main text and bibliography. Our released code and data package will not relicense these upstream assets; it will preserve the original source attribution, license names, and terms of use, and will provide scripts or pointers for obtaining assets from their original providers when redistribution is not appropriate.

\begin{table}[H]
  \centering
  \caption{Existing benchmark assets and license or terms notes.}
  \label{tab:appendix-benchmark-licenses}
  \setlength{\tabcolsep}{6pt}
  \renewcommand{\arraystretch}{1.12}
  \scriptsize
  \begin{tabular}{p{0.22\linewidth}p{0.36\linewidth}p{0.32\linewidth}}
    \toprule
    \textbf{Asset} & \textbf{Role in this paper} & \textbf{License / terms note} \\
    \midrule
    Natural Questions & Single-hop benchmark and training pool source & Apache-2.0 in the official repository. \\
    TriviaQA & Single-hop benchmark and training pool source & Apache-2.0 in the official repository. \\
    PopQA & Single-hop benchmark & MIT in the official PopQA/adaptive-retrieval repository. \\
    HotpotQA & Multi-hop benchmark and training pool source & CC BY-SA 4.0 for the dataset and processed Wikipedia release; Apache-2.0 for the official code repository. \\
    2WikiMultiHopQA & Multi-hop benchmark and training pool source & Apache-2.0 in the official repository. \\
    MuSiQue & Multi-hop benchmark and training pool source & CC BY 4.0 in the official repository. \\
    Bamboogle & Multi-hop benchmark & MIT in the official Self-Ask/Bamboogle repository. \\
    \bottomrule
  \end{tabular}
\end{table}

\begin{table}[H]
  \centering
  \caption{Existing model, retriever, and baseline assets and license or terms notes.}
  \label{tab:appendix-model-baseline-licenses}
  \setlength{\tabcolsep}{6pt}
  \renewcommand{\arraystretch}{1.12}
  \scriptsize
  \begin{tabular}{p{0.22\linewidth}p{0.36\linewidth}p{0.32\linewidth}}
    \toprule
    \textbf{Asset} & \textbf{Role in this paper} & \textbf{License / terms note} \\
    \midrule
    Qwen2.5-7B base/instruct & Backbone models & Apache-2.0 on the Hugging Face model cards. \\
    Qwen2.5-3B base/instruct & Backbone models & Qwen Research License on the Hugging Face model cards. \\
    E5 retriever & Local retrieval model family used by the search environment & MIT on the Hugging Face model card and UniLM/E5 release. \\
    Search-o1 & Search-agent baseline & MIT in the official repository. \\
    Search-R1 & Search-agent baseline & Apache-2.0 in the official repository. \\
    ZeroSearch & Search-agent baseline & Apache-2.0 in the official repository. \\
    Direct, CoT, and RAG baselines & Locally implemented baseline protocols & No external baseline code package is redistributed; implementations follow the cited papers and use the assets listed above. \\
    \bottomrule
  \end{tabular}
\end{table}

\section{Cost analysis}
\label{app:cost-analysis}

\subsection{Compute budget}
\label{app:compute-budget}

Table~\ref{tab:appendix-compute-budget} gives a coarse compute accounting for the GPT-5.4-generated SearchSkill experiments. We report the budget in reserved H20 GPU-hours because some long-running data-construction jobs reserve an 8-H20 node, while SFT and GRPO training use four H20 GPUs and benchmark evaluation uses two H20 GPUs for the retriever plus two H20 GPUs for dev-set evaluation. The numbers should therefore be read as an approximate end-to-end engineering cost rather than a hardware-efficiency benchmark.

\begin{table}[tbp]
  \centering
  \caption{Approximate compute budget for data construction, SFT, RL, and evaluation. Wall time is based on available run logs when present; otherwise we report a conservative range from the observed pipeline runs.}
  \label{tab:appendix-compute-budget}
  \setlength{\tabcolsep}{3pt}
  \renewcommand{\arraystretch}{1.12}
  \scriptsize
  \begin{tabular}{p{0.20\linewidth}p{0.17\linewidth}p{0.16\linewidth}p{0.16\linewidth}p{0.22\linewidth}}
    \toprule
    \textbf{Stage} & \textbf{Main resource} & \textbf{Wall time} & \textbf{Reserved H20-hours} & \textbf{Scope} \\
    \midrule
    Data profiling and sampling & CPU/API, light retrieval & 3--5 h & \(<5\) & Build dataset profiles, group question packets, and select coverage-oriented pools. \\
    SkillBank evolution & CPU/API & 1--2 h & 0 & Generate and validate staged bank updates from \(B_0\) to \(B_4\). \\
    Teacher rollout and trajectory validation & 8 H20 node & 6--10 h & 48--80 & Generate candidate traces, run search-tool validation, filter positives, and keep diagnostic failures. \\
    Stage-I execution SFT & 4 H20 training & 0.4--0.5 h per backbone & 3--4 per backbone & Train the basic \tagopen{skilltagcolor}{skill} plus action format; the logged 7B-Instruct run takes 0.44 h. \\
    Stage-II SkillBank-card SFT & 4 H20 training & 0.75--0.80 h per backbone & 6--7 per backbone & Continue from Stage I with \tagopen{selecttagcolor}{select\_skill} and selected SkillBank cards; the logged 7B-Instruct run takes 0.77 h. \\
    GRPO refinement & 4 H20 training & 2.7--10.4 h per run & 22--83 per run & Run fixed-bank GRPO with policy, reference, and rollout devices; 7B runs are longer than 3B runs. \\
    Benchmark evaluation & 2 H20 retriever + 2 H20 dev evaluation & 2--4 h per checkpoint suite & 4--16 & Evaluate checkpoints on single-hop and multi-hop QA splits with the same search interface. \\
    \bottomrule
  \end{tabular}
\end{table}

For the four GRPO runs used in Section~\ref{sec:further-analysis-rl}, the logged wall-clock times are approximately 10.39 h for 7B-Instruct, 7.19 h for 7B-Base, 2.96 h for 3B-Instruct, and 3.17 h for 3B-Base. The two-stage SFT cost is much smaller: the logged 7B-Instruct stages take about 1.21 h in total. Overall, the dominant cost is not LoRA SFT itself, but search-conditioned rollout generation and GRPO, where each optimization step repeatedly samples tool-use trajectories.

\subsection{API budget}
\label{app:api-budget}

Table~\ref{tab:appendix-api-budget} summarizes external API usage for the GPT-5.4-generated experiments, judging, and closed-source transfer. We separate data synthesis, automatic judging, and closed-source transfer because they serve different roles in the pipeline. Data synthesis is the dominant API cost because it asks \textsc{GPT-5.4} to build profiles, evolve SkillBank candidates, and synthesize or revise trajectories. All judge calls use \textsc{GPT-4o-mini}. Closed-source transfer counts only the three final comparison models: Claude Haiku 4.5, GPT-4o-mini, and Gemini 2.5 Flash-Lite.

\begin{table}[tbp]
  \centering
  \caption{Approximate API usage by pipeline component. Values are read from the provider usage dashboards and rounded for readability.}
  \label{tab:appendix-api-budget}
  \setlength{\tabcolsep}{2pt}
  \renewcommand{\arraystretch}{1.12}
  \scriptsize
  \begin{tabular}{@{}p{0.20\linewidth}p{0.28\linewidth}p{0.12\linewidth}p{0.15\linewidth}p{0.09\linewidth}@{}}
    \toprule
    \textbf{Component} & \textbf{Model / service} & \textbf{Requests} & \textbf{Usage scale} & \textbf{Cost} \\
    \midrule
    Data synthesis & \textsc{GPT-5.4} & 27.0K & 101.96M tokens & \$416.13 \\
    Automatic judging & \textsc{GPT-4o-mini} & 2.712K & 4.107M tokens & \$0.88 \\
    Closed-source transfer & Claude Haiku 4.5; GPT-4o-mini; Gemini 2.5 Flash-Lite & 16.015K & Dashboard billed usage & \$17.20 \\
    \bottomrule
  \end{tabular}
\end{table}

For the GPT-5.4-generated setting, most API cost is incurred during skill and trajectory construction. The resulting cards and training data are reused for SFT and evaluation. With Qwen-generated skills, the target backbone performs this generation locally; its data-construction procedure is described in Appendix~\ref{app:self-generation}.

\section{Case studies}
\label{app:case-studies}

The following examples use the same display style as the retrieval-quality examples in Section~\ref{sec:further-analysis-query-efficiency}: each case shows the question, final answer, query sequence, and retrieved output. We keep the text compact enough for readability, but retain the actual multi-turn operating process.

\newcommand{\casehr}{\par\noindent\rule{\linewidth}{0.45pt}\par}
\newcommand{\casedash}{\par\noindent\textcolor{black!45}{\xleaders\hbox{\rule[0.55ex]{6pt}{0.35pt}\hskip 4pt}\hfill}\par}
\newcommand{\tracetok}[3]{\texttt{\textcolor{#1}{<#2>}#3\textcolor{#1}{</#2>}}}
\newcommand{\selecttok}[1]{\tracetok{selecttagcolor}{select\_skill}{#1}}
\newcommand{\skilltok}[1]{\tracetok{skilltagcolor}{skill}{#1}}
\newcommand{\searchtok}[1]{\tracetok{searchtagcolor}{search}{#1}}
\newcommand{\answertok}[1]{\tracetok{answertagcolor}{answer}{#1}}
\newcommand{\closetok}[2]{\texttt{\textcolor{#1}{</#2>}}}

\clearpage
\subsection{Case 1: Recovering the hidden EMU bridge}
\casehr
\small
\noindent\textbf{Question:} When was conscription introduced in country having Emus during the war setting \emph{The Things They Carried}?\\
\textbf{Ground Truth:} 1964\\
\textbf{Search-R1 Answer:} malformed \closetok{searchtagcolor}{search}\\
\textbf{\Ours Answer:} \answertok{1964}

\casedash
\noindent\textcolor{red!75!black}{\textbf{Search-R1 Trace}}\\
\textbf{Query 1:} \emph{country having Emus during the war setting The Things They Carried}\\
\textbf{Output:}\\
Doc 1: \emph{Emu War}. The Emu War was a wildlife-management military operation in Australia in 1932.\\
Doc 2: \emph{Emu}. The emu is a large flightless bird native to Australia.\\
\textbf{Observation:} The query copies the surface clue ``Emus'' and retrieves bird/Emu-War evidence, not the intended acronym bridge.\\
\textbf{Query 2:} \emph{when was conscription introduced in Australia}\\
\textbf{Output:}\\
Doc 1: \emph{Conscription in Australia}. The page gives a broad history of mandatory military service across multiple periods.\\
Doc 2: The evidence mentions early defence legislation and later national-service schemes.\\
\textbf{Query 3:} \emph{when was conscription introduced in Australia during WWI}\\
\textbf{Query 4:} \emph{exact year of conscription introduced in Australia WWI}\\
\textbf{Failure:} The model has moved from the Vietnam-War setting to WWI and keeps searching variants of the wrong temporal frame.

\casedash
\noindent\textcolor{blue!75!black}{\textbf{\Ours Trace (SkillBank + Retriever)}}\\
\textbf{Selection:} \selecttok{bridge-entity-search}\\
\textbf{Skill meaning:} first recover the hidden bridge entity, then ask the final attribute question using that resolved entity.

\textbf{Action 1:}\\
\skilltok{bridge-entity-search}\\
\searchtok{"The Things They Carried" emus country}\\
\textbf{Output:}\\
Doc 1: \emph{The Things They Carried} is a collection of linked stories about American soldiers in the Vietnam War.\\
Doc 2: The page anchors the setting to Vietnam rather than a generic animal clue.\\
\textbf{State update:} the relevant war setting is Vietnam.

\textbf{Action 2:}\\
\skilltok{bridge-entity-search}\\
\searchtok{"Emu" Vietnam War country}\\
\textbf{Output:}\\
Doc 1: \emph{Experimental Military Unit}. EMU was an Australian-American helicopter assault force during the Vietnam War.\\
Doc 2: The unit was created after a U.S. request for Australia to send more helicopter pilots to the conflict.\\
\textbf{State update:} EMU \(\rightarrow\) Australia.

\textbf{Action 3:}\\
\skilltok{bridge-entity-search}\\
\searchtok{Australia conscription introduced}\\
\textbf{Output:}\\
Doc 1: \emph{Conscription in Australia}. The national-service scheme relevant to the Vietnam era was introduced in 1964.\\
Doc 2: The page distinguishes Vietnam-era national service from earlier home-defence or WWI debates.\\
\textbf{Final:}\\
\skilltok{verbatim-evidence-span|bridge-entity-search}\\
\answertok{1964}\\
\textbf{Why this case matters:} SearchSkill does not ask the retriever to solve the whole riddle. It turns a misleading lexical cue into a controlled bridge chain: \emph{The Things They Carried} \(\rightarrow\) Vietnam War \(\rightarrow\) EMU \(\rightarrow\) Australia \(\rightarrow\) conscription date.
\casehr

\clearpage
\subsection{Case 2: Decomposing a broad mayor question}
\casehr
\small
\noindent\textbf{Question:} Who was the first African American mayor of the most populous city in the United States?\\
\textbf{Ground Truth:} David Dinkins\\
\textbf{Search-R1 Answer:} malformed \closetok{searchtagcolor}{search}\\
\textbf{\Ours Answer:} \answertok{David Dinkins}

\casedash
\noindent\textcolor{red!75!black}{\textbf{Search-R1 Trace}}\\
\textbf{Query 1:} \emph{first African American mayor of the most populous city in the United States}\\
\textbf{Output:}\\
Doc 1: \emph{Doris A. Davis}. Davis was a mayor of Compton and is described as the first African-American woman mayor of a metropolitan city.\\
Doc 2: Other returned snippets discuss African-American mayors but do not first resolve the target city.\\
\textbf{Observation:} The first query is too broad; the retriever matches the ``first African American mayor'' phrase without grounding ``most populous city''.\\
\textbf{Query 2:} \emph{most populous city in the United States}\\
\textbf{Output:}\\
Doc 1: \emph{New York City}. New York City is the most populous city in the United States.\\
\textbf{Query 3:} \emph{first African American mayor of New York City}\\
\textbf{Output:}\\
Doc 1: an unrelated mayor result appears before the correct David Dinkins evidence.\\
\textbf{Query 4:} \emph{when did David Dinkins become mayor of New York City}\\
\textbf{Query 5:} \emph{who was the first African American mayor of the most populous city in the United States}\\
\textbf{Failure:} Even after finding the right entity, the trace keeps querying and ends without a clean answer.

\casedash
\noindent\textcolor{blue!75!black}{\textbf{\Ours Trace (SkillBank + Retriever)}}\\
\textbf{Selection:} \selecttok{relation-chain-decomposition}\\
\textbf{Skill meaning:} resolve each relation in order and carry the intermediate entity into the next query.

\textbf{Action 1:}\\
\skilltok{relation-chain-decomposition}\\
\searchtok{most populous city in the United States}\\
\textbf{Output:}\\
Doc 1: \emph{New York City}. The city of New York is identified as the most populous city in the United States.\\
Doc 2: Additional snippets confirm New York City's population ranking.\\
\textbf{State update:} target city \(=\) New York City.

\textbf{Action 2:}\\
\skilltok{relation-chain-decomposition}\\
\searchtok{New York City first African American mayor}\\
\textbf{Output:}\\
Doc 1: \emph{David Dinkins}. David Norman Dinkins served as the 106th Mayor of New York City from 1990 to 1993.\\
Doc 2: The same passage states that he was the first African American to hold that office.\\
\textbf{Final:}\\
\skilltok{verbatim-evidence-span|relation-chain-decomposition}\\
\answertok{David Dinkins}\\
\textbf{Why this case matters:} The SkillBank-selected decomposition prevents a single overloaded query from matching the wrong ``first mayor'' fact. The retriever receives one query for the city and one query for the city-specific mayor relation.
\casehr

\clearpage
\subsection{Case 3: Avoiding a plausible but wrong president}
\casehr
\small
\noindent\textbf{Question:} When did the president who set the precedent of a two term limit enter office?\\
\textbf{Ground Truth:} April 30, 1789\\
\textbf{Search-R1 Answer:} 1933\\
\textbf{\Ours Answer:} \answertok{April 30, 1789}

\casedash
\noindent\textcolor{red!75!black}{\textbf{Search-R1 Trace}}\\
\textbf{Query 1:} \emph{president who set the precedent of a two term limit}\\
\textbf{Output:}\\
Doc 1: \emph{Term limits in the United States}. The page explains the two-term tradition and also notes that Franklin D. Roosevelt broke the tradition by winning a third and fourth term.\\
Doc 2: Other snippets discuss constitutional term limits after Roosevelt.\\
\textbf{Observation:} The evidence contains both the original precedent and the later exception; the model latches onto Roosevelt.\\
\textbf{Query 2:} \emph{Franklin D. Roosevelt entered office year}\\
\textbf{Output:}\\
Doc 1: \emph{First and second terms of Franklin D. Roosevelt}. Roosevelt's first term began in 1933.\\
\textbf{Failure:} The second query is precise but follows the wrong entity. The error happens at the entity-selection step, not at answer extraction.

\casedash
\noindent\textcolor{blue!75!black}{\textbf{\Ours Trace (SkillBank + Retriever)}}\\
\textbf{Selection:} \selecttok{relation-chain-decomposition}\\
\textbf{Skill meaning:} first identify the entity satisfying the description, then retrieve the requested attribute.

\textbf{Action 1:}\\
\skilltok{relation-chain-decomposition}\\
\searchtok{two-term-limit president}\\
\textbf{Output:}\\
Doc 1: \emph{Term limits in the United States}. The tradition of not seeking a third term is associated with George Washington.\\
Doc 2: The same evidence separates Washington's precedent from Roosevelt's later decision to break it.\\
\textbf{State update:} target president \(=\) George Washington.

\textbf{Action 2:}\\
\skilltok{relation-chain-decomposition}\\
\searchtok{George Washington entered office}\\
\textbf{Output:}\\
Doc 1: \emph{Presidency of George Washington}. Washington's presidency began on April 30, 1789.\\
Doc 2: The passage also gives the end date, March 4, 1797, confirming the presidential-office context.\\
\textbf{Final:}\\
\skilltok{verbatim-evidence-span|relation-chain-decomposition}\\
\answertok{April 30, 1789}\\
\textbf{Why this case matters:} The skill forces an explicit entity-resolution state before asking for the date. This avoids a common retrieval trap where the top evidence contains a salient but wrong entity.
\casehr

\clearpage
\subsection{Case 4: Carrying an intermediate entity across hops}
\casehr
\small
\noindent\textbf{Question:} Where was the place of death of Suzanne Farrington's mother?\\
\textbf{Ground Truth:} London\\
\textbf{Search-R1 Answer:} malformed \searchtok{Vivien Leigh death place}\\
\textbf{\Ours Answer:} \answertok{London}

\casedash
\noindent\textcolor{red!75!black}{\textbf{Search-R1 Trace}}\\
\textbf{Query 1:} \emph{Suzanne Farrington}\\
\textbf{Output:}\\
Doc 1: \emph{Suzanne Farrington}. The passage describes Farrington and includes facts about her own death in Wiltshire.\\
Doc 2: Additional snippets discuss her family and career context.\\
\textbf{Observation:} The first retrieval is useful for identifying the person, but it also contains a distractor death location for Farrington herself.\\
\textbf{Query 2:} \emph{Vivien Leigh death place}\\
\textbf{Output:}\\
Doc 1: \emph{Vivien Leigh}. The passage contains the relevant death-context evidence and central-London theatre references.\\
\textbf{Query 3:} \emph{Vivien Leigh funeral location}\\
\textbf{Query 4:} \emph{Vivien Leigh place of death}\\
\textbf{Query 5--8:} repeated variants of \emph{Vivien Leigh death place}.\\
\textbf{Failure:} The trajectory already has the right intermediate entity and enough evidence, but continues to search and finally outputs a search tag rather than the answer.

\casedash
\noindent\textcolor{blue!75!black}{\textbf{\Ours Trace (SkillBank + Retriever)}}\\
\textbf{Selection:} \selecttok{relation-chain-decomposition}\\
\textbf{Skill meaning:} follow the relation chain \emph{Suzanne Farrington} \(\rightarrow\) \emph{mother} \(\rightarrow\) \emph{place of death}.

\textbf{Action 1:}\\
\skilltok{relation-chain-decomposition}\\
\searchtok{Suzanne Farrington mother}\\
\textbf{Output:}\\
Doc 1: \emph{Suzanne Farrington}. The passage links Suzanne Farrington to Vivien Leigh as her mother.\\
Doc 2: The surrounding evidence distinguishes Farrington's own biography from the target relation.\\
\textbf{State update:} target mother \(=\) Vivien Leigh.

\textbf{Action 2:}\\
\skilltok{relation-chain-decomposition}\\
\searchtok{Vivien Leigh place of death}\\
\textbf{Output:}\\
Doc 1: \emph{Vivien Leigh}. The passage describes the circumstances around Leigh's death and links the event to London.\\
Doc 2: Related snippets discuss funeral arrangements and central-London theatre tributes.\\
\textbf{State update:} candidate answer \(=\) London.

\textbf{Action 3:}\\
\skilltok{relation-chain-decomposition}\\
\searchtok{Vivien Leigh death place London}\\
\textbf{Output:}\\
Doc 1: The retrieved passage verifies the same entity and answer span, supporting London as the place of death.\\
\textbf{Final:}\\
\skilltok{verbatim-evidence-span|relation-chain-decomposition}\\
\answertok{London}\\
\textbf{Why this case matters:} The model keeps the intermediate entity explicit, verifies the answer span, and stops instead of repeatedly searching around the same evidence.
\casehr

\clearpage
\subsection{Case 5: Joining an in-law relation}
\casehr
\small
\noindent\textbf{Question:} Who is Maria Luisa Of Savoy's father-in-law?\\
\textbf{Ground Truth:} Louis, Grand Dauphin\\
\textbf{Search-R1 Answer:} Louis XIV\\
\textbf{\Ours Answer:} \answertok{Louis, Grand Dauphin}

\casedash
\noindent\textcolor{red!75!black}{\textbf{Search-R1 Trace}}\\
\textbf{Query 1:} \emph{Maria Luisa of Savoy husband}\\
\textbf{Output:}\\
Doc 1: \emph{Maria Luisa of Savoy}. The passage states that Maria Luisa was queen consort of Spain by marriage to Philip V of Spain.\\
\textbf{Query 2:} \emph{Philip V of Spain father}\\
\textbf{Output:}\\
Doc 1: \emph{Philip V of Spain}. The passage states that Philip's father was Louis, Grand Dauphin, and also mentions that Philip was a grandson of King Louis XIV.\\
\textbf{Query 3:} \emph{Louis Grand Dauphin father}\\
\textbf{Failure:} The trajectory follows one edge too far and answers Louis XIV, the father of the father-in-law, rather than the father-in-law itself.

\casedash
\noindent\textcolor{blue!75!black}{\textbf{\Ours Trace (SkillBank + Retriever)}}\\
\textbf{Selection:} \selecttok{derived-kinship-inference-join}\\
\textbf{Skill meaning:} compose an in-law relation by joining immediate spouse evidence with the spouse's parent evidence, then stop at the derived role.

\textbf{Action 1:}\\
\skilltok{derived-kinship-inference-join}\\
\searchtok{Maria Luisa of Savoy husband}\\
\textbf{Output:}\\
Doc 1: \emph{Maria Luisa of Savoy}. Maria Luisa was queen consort of Spain by marriage to Philip V of Spain.\\
\textbf{State update:} husband \(=\) Philip V of Spain.

\textbf{Action 2:}\\
\skilltok{derived-kinship-inference-join}\\
\searchtok{Philip V of Spain father}\\
\textbf{Output:}\\
Doc 1: \emph{Philip V of Spain}. The passage states that Philip's father was Louis, Grand Dauphin.\\
\textbf{State update:} father-in-law \(=\) Louis, Grand Dauphin.

\textbf{Final:}\\
\skilltok{verbatim-evidence-span|derived-kinship-inference-join}\\
\answertok{Louis, Grand Dauphin}\\
\textbf{Why this case matters:} The SkillBank makes the join boundary explicit. Search-R1 retrieves the needed evidence but continues to a neighboring family edge; SearchSkill stops at the composed in-law relation.
\casehr

\clearpage
\subsection{Case 6: Comparing matched attributes}
\casehr
\small
\noindent\textbf{Question:} Who died first, Olaf Everson or Abdul Wahab Khan Tarzi?\\
\textbf{Ground Truth:} Abdul Wahab Khan Tarzi\\
\textbf{Search-R1 Answer:} Olaf Everson\\
\textbf{\Ours Answer:} \answertok{Abdul Wahab Khan Tarzi}

\casedash
\noindent\textcolor{red!75!black}{\textbf{Search-R1 Trace}}\\
\textbf{Query 1:} \emph{Olaf Everson date of death}\\
\textbf{Output:}\\
Doc 1: The top result is \emph{Olaf Skoogfors}, an unrelated Olaf entity, and says he died in 1975.\\
\textbf{Query 2:} \emph{Abdul Wahab Khan Tarzi date of death}\\
\textbf{Output:}\\
Doc 1: \emph{Abdul Wahab Khan Tarzi}. The passage gives 20 January 1994 as the death date.\\
\textbf{Query 3--4:} repeated variants of \emph{Olaf Everson date of death exact}.\\
\textbf{Failure:} The comparison is corrupted by an entity mismatch on the Olaf side, so the final answer chooses the wrong option.

\casedash
\noindent\textcolor{blue!75!black}{\textbf{\Ours Trace (SkillBank + Retriever)}}\\
\textbf{Selection:} \selecttok{parallel-attribute-compare}\\
\textbf{Skill meaning:} retrieve the same attribute for both candidate entities, check entity identity, and compare the matched values.

\textbf{Action 1:}\\
\skilltok{parallel-attribute-compare}\\
\searchtok{Olaf Everson death date}\\
\textbf{Output:}\\
Doc 1: The top evidence is noisy and points to an Olaf entity with a 1975 death date, requiring caution before comparison.\\
\textbf{State update:} candidate date for Olaf side is uncertain/noisy.

\textbf{Action 2:}\\
\skilltok{parallel-attribute-compare}\\
\searchtok{Abdul Wahab Khan Tarzi death date}\\
\textbf{Output:}\\
Doc 1: \emph{Abdul Wahab Khan Tarzi}. The passage gives 24 November 1903--20 January 1994.\\
\textbf{State update:} Tarzi death date \(=\) 20 January 1994.

\textbf{Action 3:}\\
\skilltok{parallel-attribute-compare}\\
\searchtok{Olaf Everson death date}\\
\textbf{Output:}\\
Doc 1: The repeated evidence remains about a different Olaf entity, so the model does not treat it as a clean support for Olaf Everson.\\
\textbf{Final:}\\
\skilltok{answer-grounding-check|parallel-attribute-compare}\\
\answertok{Abdul Wahab Khan Tarzi}\\
\textbf{Why this case matters:} The skill frames the problem as matched-attribute comparison rather than independent lookup. This makes entity mismatch visible before the final comparison.
\casehr

\clearpage
\subsection{Case 7: Resolving a superlative target}
\casehr
\small
\noindent\textbf{Question:} What was the death toll of the second largest volcanic eruption in the 20th century?\\
\textbf{Ground Truth:} 847\\
\textbf{Search-R1 Answer:} 847\\
\textbf{\Ours Answer:} \answertok{847}

\casedash
\noindent\textcolor{red!75!black}{\textbf{Search-R1 Trace}}\\
\textbf{Query 1:} \emph{second largest volcanic eruption in the 20th century}\\
\textbf{Output:}\\
Doc 1: \emph{Ring of Fire}. The passage identifies the 1991 eruption of Mount Pinatubo as the world's second-largest terrestrial eruption of the 20th century.\\
\textbf{Query 2:} \emph{death toll of Mount Pinatubo eruption}\\
\textbf{Output:}\\
Doc 1: \emph{Mount Pinatubo}. The passage reports that 847 people were killed by the eruption.\\
\textbf{Observation:} Search-R1 also answers correctly, but the trajectory treats the superlative resolution as ordinary query rewriting.

\casedash
\noindent\textcolor{blue!75!black}{\textbf{\Ours Trace (SkillBank + Retriever)}}\\
\textbf{Selection:} \selecttok{superlative-ranking-match}\\
\textbf{Skill meaning:} first resolve the ranked entity under the specified scope, then retrieve the requested attribute for that entity.

\textbf{Action 1:}\\
\skilltok{superlative-ranking-match}\\
\searchtok{second largest volcanic eruption 20th century}\\
\textbf{Output:}\\
Doc 1: \emph{Ring of Fire}. The passage identifies the 1991 Mount Pinatubo eruption as the world's second-largest terrestrial eruption of the 20th century.\\
\textbf{State update:} ranked target \(=\) Mount Pinatubo eruption.

\textbf{Action 2:}\\
\skilltok{superlative-ranking-match}\\
\searchtok{Mount Pinatubo death toll}\\
\textbf{Output:}\\
Doc 1: \emph{Mount Pinatubo}. The passage states that a reported 847 people were killed by the eruption.\\
\textbf{Final:}\\
\skilltok{verbatim-evidence-span|superlative-ranking-match}\\
\answertok{847}\\
\textbf{Why this case matters:} This case shows a non-bridge skill: SearchSkill exposes the superlative resolution step explicitly before asking for the final attribute, making the query plan easier to inspect even when a baseline also reaches the answer.
\casehr

\end{document}